\documentclass[journal]{IEEEtran}
\ifCLASSINFOpdf
\else
\fi

\usepackage{graphics}
\usepackage{url}

\usepackage{bm}
\usepackage{float}
\usepackage{graphicx}
\usepackage{subfigure}
\usepackage{amsmath,epsfig,algorithm,algorithmic,multirow,xcolor,amsmath}

\begin{document}
%
\title{Unsupervised Band Selection of Hyperspectral Images via Multi-dictionary Sparse Representation}
%
%
%

\author{Fei~Li,
        Pingping~Zhang
        and Huchuan~Lu
\thanks{
Copyright (c) 2018 IEEE. Personal use of this material is permitted. However, permission to use this material for any other purposes must  be obtained from the IEEE by sending an email to \textcolor{blue}{\underline{pubs-permissions@ieee.org}}.

All of the authors are with the School of Information and Communication Engineering, Faculty of Electronic Information and Electrical Engineering, Dalian University of Technology, Dalian 116024, China.
F. Li is also with the Engineering Training Center, Shenyang Aerospace University, Shenyang 110136, China.
PP. Zhang is currently visiting the University of Adelaide, supported by the China Scholarship Council (CSC) program.
The corresponding author is Prof. Huchuan Lu.
Email: alinafeili@mail.dlut.edu.cn; jssxzhpp@mail.dlut.edu.cn; lhchuan@dlut.edu.cn.

This work is supported in part by the National Natural Science Foundation of China (NNSFC), No. 61502070, No. 61528101 and No. 61472060.
}}
\markboth{Submitted to IEEE Transactions on Circuits and Systems for Video Technology,~Vol.~XX, No.~XX, Feb~2018}
{Shell \MakeLowercase{\textit{et al.}}: Bare Demo of IEEEtran.cls for IEEE Journals}
%



\maketitle

\begin{abstract}
Hyperspectral images have far more spectral bands than ordinary multispectral images.
Rich band information provides more favorable conditions for the tremendous applications.
However, significant increase in the dimensionality of spectral bands may lead to the curse of dimensionality, especially for classification applications.
Furthermore, there are a large amount of redundant information among the raw image cubes due to water absorptions, sensor noises and other influence factors.
Band selection is a direct and effective method to remove redundant information and reduce the spectral dimension
for decreasing computational complexity and avoiding the curse of dimensionality.
In this paper, we present a novel learning framework for band selection based on the idea of sparse representation.
More specifically, first each band is approximately represented by the linear combination of other bands, then the original band image can be represented by a multi-dictionary learning mechanism.
As a result, a group of weights can be obtained by sparse optimization for all bands.
Finally, the specific bands will be selected, if they get higher weights than other bands in the representation of the original image.
Experimental results on three widely used hyperspectral datasets show that our proposed algorithm achieves better performance in hyperspectral image classification, when compared with other state-of-art band selection methods.
\end{abstract}


%
\IEEEpeerreviewmaketitle

\section{Introduction}
%
%
%
%
\IEEEPARstart{H}{yperspectral} remote sensing is a critical research area for remote sensing applications~\cite{Chang2007Hyperspectral}.
Recently, the development of hyperspectral remote sensing is accelerated with the enhancing spectral resolution of remote sensors, more insights about spectral band features of ground objects and the discovery on numerous ground objects' characteristics hidden in the narrow spectrum range.

Generally, hyperspectral images (HSI) of remote sensing are acquired by using airborne or spaceborne hyperspectral remote sensors.
The number of spectral bands of hyperspectral images usually reaches hundreds or even thousands~\cite{Chang2003Hyperspectral}.
This rich spectral information enhances and expands the potential application of hyperspectral images.
For example, hyperspectral images have been successfully used in surface vegetation monitoring, geological mapping and mineral exploration~\cite{ͯÇìϲ,Schaepman2009Earth}.
However, the huge volume data in hyperspectral images also brings many obstacles in data transmission, data storage, etc.
In particular, the application of hyperspectral images faces following important issues.
First, with the increase of the volume data, the computational complexity becomes higher and higher.
Second, the increase of spectral dimension can easily lead to the curse of dimensionality or Hughes phenomenon in the remote sensing applications~\cite{hughes1968mean,RichardErnest1961Adaptive}.
In addition, there is redundant information in hyperspectral image cubes due to the spectral reflectance of most materials changes only gradually over certain spectral regions and many contiguous bands are highly correlated~\cite{Jia1994Efficient}.
More importantly, the complete data with hundreds of spectral bands is not designed for any particular problem, only providing opportunities for a wide range of applications.
Not all spectral bands are useful and necessary in a given problem.
In another word, a given spectral band may be a useful feature in a problem, but not for others~\cite{jia2013feature}.
Therefore, it is necessary to select and reduce the spectral dimension for efficient hyperspectral image applications.

In general, there are two kinds of approaches to reduce the spectral dimension.
One is robust feature extraction and the other is specific band selection (also called feature selection).
The former usually generates a low-dimensional spectral data by using a transformation matrix based on a certain criteria.
For instance, the transformation criterion can be principal component analysis (PCA)~\cite{Zubko2007Principal}, linear discriminant analysis (LDA)~\cite{Bandos2009Classification}, non-parametric weighted feature extraction (NWFE)~\cite{kuo2004nonparametric}, and so on.
However, aforementioned methods are classical unsupervised or supervised feature extraction methods, and they inevitably change the physical meaning of the original bands.
The latter is to identify the best subset of bands from all the original bands based on an adaptive selection criterion~\cite{jia2013feature}.
The band selection method can preserve the original meaning while reduce the spectral dimension.
In this paper we focus on the band selection methods.
%
According to the need of prior knowledge, existing band selection methods can be roughly classified into two categories, \emph{i.e.}, supervised band selection and unsupervised band selection.
Supervised band selection methods require rich prior knowledge about pixels' labels~\cite{wang2007novel,du2008similarity}.
For example, canonical analysis~\cite{chang1999joint}, Jeffries-Matusita distance and its extensions~\cite{bruzzone1995extension}, contract divergence~\cite{huang2005band}, cluster space separability~\cite{jia2002cluster} were adopted as the selection criteria.
Though effective, there are some limitations in these supervised approaches because they usually have some specific purposes and are applied in some specific situations.
%
%
Unlike supervised methods, unsupervised band selection methods do not need any prior knowledge about pixel label information, and they are more practical than supervised methods.
Recently, many researchers have made a lot of efforts on unsupervised band selection, and put forward several useful methods.
For example, Sui et al.~\cite{sui2014unsupervised,sui2015unsupervised} proposed an unsupervised band selection method which integrates the overall accuracy and redundancy into the band selection process.
Through an optimization model, a balance parameter is designed to trade off the overall accuracy and redundancy.
Du et al.~\cite{du2008similarity} proposed an unsupervised band selection method based on the similarity of bands.
In~\cite{yang2011efficient}, the dissimilarity of bands is also utilized for the specific band selection.
In addition, parallel implementations were adopt to accelerate the speed of band selection methods~\cite{yang2011unsupervised}.

Contrary to previous works, in this paper we propose a novel unsupervised band selection algorithm for hyperspectral image processing that utilizes the sparsity of the input sample vector.
Our proposed algorithm is based on an effective sparse representation model in which the spatial image of a band is approximately represented by linear combinations of a few spatial image group (called atoms) from the entire hyperspectral image.
The sparse vector, representing the atoms and associated weights for the input sample, can be obtained by solving an optimization problem constrained by the sparsity level and reconstruction accuracy.
The specific bands can be determined by the value of associated weights which are not zeros under multiple given overcomplete dictionaries.
We sort bands in descending order by the value of non-zero weights, and the first several bands are selected as the final reduced representation.
In order to evaluate the performance of the proposed method, extensive comparisons against several unsupervised band selection methods are presented.
These compared methods are well chosen to cover recent tendencies in this field.
Experimental results on three widely used hyperspectral image classification datasets show that our proposed algorithm achieves superior performance and significantly outperforms other state-of-the-art unsupervised band selection methods.
%

The rest of this paper is organized as follows.
In Section II, we introduce the proposed unsupervised band selection method.
Then we present the experimental results on some well known hyperspectral images in Section III.
Finally, we make conclusions in Section IV.
\begin{figure*}[!htb]
\centering
\includegraphics[width=0.95\textwidth]{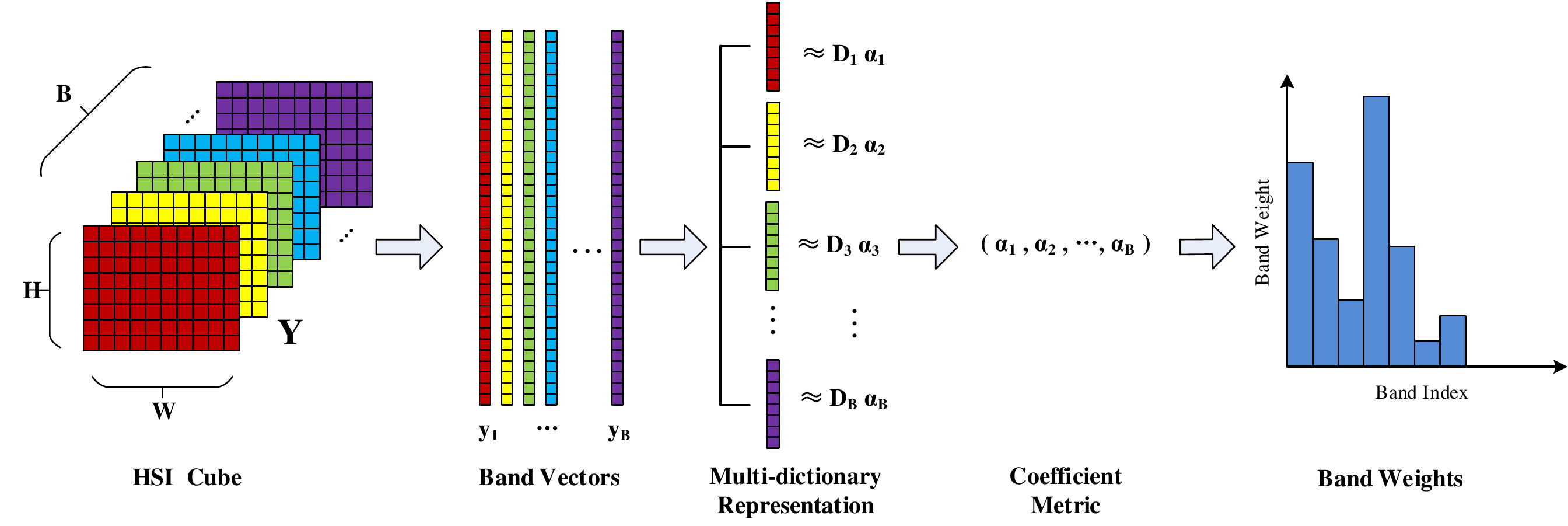}
\caption{The schematic pipeline of the proposed band selection method. First, each band of the three-dimension hyperspectral data \textbf{Y} is stretched into a high-dimension vector. Then, the dimension of the vector is reduced by random sampling, so that the dimension of the vector is far less than the number of bands. Next, multi-dictionary learning is applied to the band vectors, and each band is approximately represented by the linear combination of other bands. Finally, the weight of each band is calculated by the sparse coefficient. The first several bands with large weights are selected as the final representation.}
\label{fig:res}
\end{figure*}
\section{Sparse representation based band selection}
In recent years, sparse representation comes to be one of the hotspots in digital signal processing.
In brief, sparse representation is the decomposition process of the original signal, and in this decomposition process the input signal will be represented as a linear approximation of the dictionary~\cite{Park2010Sparse}.
In the field of image processing, sparse representation has been successfully applied in image denoising~\cite{Elad2006Image}, image restoration~\cite{Mairal2008Sparse,Zhang2014Group}, image recognition~\cite{Wagner2012Toward,Lu2013Face} and so on.
In the same way, the sparse representation method has also achieved great success in hyperspectral image processing, such as classification and target detection.
In this section, we first introduce sparse representation based image processing algorithm that represents the sample through a sparse linear combination of samples from a dictionary.
We then presents the proposed multi-dictionary sparse representation method to solve the band selection problem.

\subsection{Sparse Representation}
The sparsity of signals has become a particularly powerful prior in many signal processing tasks, especially in the area of computer vision and pattern recognition~\cite{wright2010sparse}.
Recently, sparse representation methods have been extended to the field of hyperspectral image processing, such as hyperspectral image classification~\cite{chen2011hyperspectral,chen2013hyperspectral,he2012sparse} and hyperspectral object detection~\cite{chen2011sparse}.
In methods of hyperspectral image classification, pixels are the performed objects which can be represented by a dictionary or basis with specific sparse constraints.
A few coefficients derived from the representation can carry important information of the pixels, and can be treated as the feature of pixels.
At last, a pixel can be signed a label by searching the minimum reconstruction error through the coefficients and dictionary elements.
Formally, for the hyperspectral images classification problem, the sparse representation model can be casted as the following equation,
\begin{equation}\label{1}
\begin{aligned}
x&\approx A^{1}\alpha ^{1}+A^{2}\alpha ^{2}+...+A^{M}\alpha ^{M}\\
&=\underbrace{\begin{bmatrix}
A^{1} & ... &A^{M}
\end{bmatrix}}_\textbf{A}\underbrace{\begin{bmatrix}
\alpha ^{1}\\ ...
\\ \alpha ^{M}
\end{bmatrix}}_\alpha=\textbf{A}\alpha,
\end{aligned}
\end{equation}
where $x$ is a pixel in a hyperspectral image, $\textbf{A}$ is the dictionary including $M$ sub-dictionaries. $\alpha$ is the coefficient vector of this linear combination.
With the selected sub-dictionaries as inputs, naive classifiers, like random forest~\cite{M2005Random}, SVM~\cite{Chang2011LIBSVM}, Bayesian network~\cite{Goldszmidt1997Bayesian}, can be used for further classification.
\subsection{Sparse Representation model for band selection}
The main purpose of band selection is to find an optimal or suboptimal band subset instead of the original hyperspectral image, which can be used in sequential applications.
In another word, the subset of bands is the set that can approximately represent the original bands under some measures or the subset of bands is the collection of bands which mainly contribute to the whole hyperspectral image.
Hence, we should find out the contribution of each band to the whole image, and then choose the band according to its contribution.
Sparse representation is an effective method to rank the contribution.
When a band image is approximated by a linear combination of a dictionary, which consists of other band images, the sparse coefficients or weights will represent the contribution of each dictionary atom to the target band image.
If the weight is large, the band will make great contribution to the target band, while if the weight is small, the band will make little contribution to the target band.
We calculate sparse representation for each band by corresponding dictionaries and get a series of weights.
The contribution of each band to the whole image can be obtained by statistical weights.
As a result, the bands with larger weights are the selected bands.

When we describe the band selection problem of hyperspectral images, the spatial image of each band becomes the performed object.
Figure 1 illustrates the schematic pipeline of the proposed band selection method.
Let $\bm{Y}\in R^{W \times H \times B}$ be the original hyperspectral image.
The spatial image of every band is stretched into a real-value vector instead of a two-dimensional matrix, thus we get
\begin{equation}\label{1}
\bm{Y}=[\bm{y}_1, \bm{y}_2,...,\bm{y}_B],
\end{equation}
where $\bm{y}_i \in R^L$ ($i = 1, 2,..., B$) is the image vector of the $i$-th band, $B$ is the number of spectral bands, $L= W\times H$ is the number of pixels in the image, and $W$ and $H$ are the width and height of the band image, respectively.
The purpose of band selection is to select a best subset from the original band set.
The resulting number of bands in that subset is less than $B$.
If there is a best subset which can be found from original band set, the subset can approximately represent the original band set.
We can use the following equation to present the relation,
\begin{equation}\label{1}
    \bm{y}_i\approx \bm{D}_i\alpha_i, \bm{D}_i \subset \bm{Y},
\end{equation}
where $\bm{D}_i$ is the band subset which is exclusive of $\bm{y}_i$ and $\alpha_i$ is the linear combination coefficient for the $i$-th image vector on the subset $\bm{D}_i$.
It should be noted that there is a trivial solution for the above equation.
In the whole band set $\bm{Y}$, there is a column vector, \emph{e.g.}, $\bm{y}_i$, therefore, one can use the whole band set $\bm{Y}$ as the basis, leading to the coefficient $\alpha_i$ is one, and others are zeros.
As a result, the coefficient vector $\bm{\alpha}$ will be an identity vector which lead to same weights for all band vectors.
In order to avoid the case, we assign each band vector to a band subset $\bm{D}_i$ (also called sub-dictionary).
We will use the new sub-dictionaries $\bm{D}_i$ hereafter.
In addition, if we describe all bands by the defined band subset, the above equation can be represented as
\begin{equation}\label{1}
    \bm{Y}\approx \bm{D}\bm{\alpha},
\end{equation}
where $\bm{D}$ is the set $\{\bm{D}_i\} (i= 1, 2..., B)$ and $\bm{\alpha}$ is the set of coefficient weight $\alpha_i$.
%

According to the previous expressions, we can convert every represented band vector as
\begin{equation}\label{1}
    \bm{y}_i=\bm{D}_i\alpha_i+\bm{\beta}_i,
  \end{equation}
where $\bm{\beta}_i$ is the approximation error vector.
By evaluating the value of $\alpha_i$, we can determine the contribution of each band in $\bm{D}_i$ to the target image vector $\bm{y}_i$.
With sparse constraints on $\bm{\alpha}$, which means that if $\bm{y}_i$ is independent with the $j$-th column of $\bm{D}$, the value of $j$-th element in $\bm{\alpha}$ is zero, we can select the most important bands based on the approximation error.
If we calculate the sparse representation of all bands in $\bm{Y}$, a sparse coefficient matrix will be obtained.
Each band vector's weight is the sum of the corresponding row in the coefficient matrix.
The best $\alpha_i$ value can be found by the following optimization problem.
We will elaborate it in detail.
\subsection{Multi-dictionary Learning}
To find the optimal linear combination coefficients, we propose a novel multi-dictionary learning method.
Formally, the sparse coefficient $\alpha_i$ can be derived by solving the constrained optimization problem,
\begin{equation}\label{1}
\begin{aligned}
  \hat{\bm{\alpha}}=\text{arg min}\|\bm{\alpha}\|_0 \\
   s.t \quad \bm{y}_i=\bm{D}_i\alpha_i+\bm{\beta}_i
\end{aligned}
\end{equation}
where $||*||_{0}$ is the $L_0$ norm, which is the total number of non-zero elements in a vector.
Because the approximation error is usually restricted to a controllable interval in empirical data, the constraint in previous equation can be relaxed to the inequality form,
\begin{equation}\label{1}
\begin{aligned}
  \hat{\bm{\alpha}}=\text{arg min}\|\bm{\alpha}\|_0\\
  s.t\quad \|\bm{D}_i\alpha_i-\bm{y}_i\|_2 \leq\sigma
\end{aligned}
\end{equation}
where $\sigma$ is the error tolerance.
Thus, the above optimization problem can be further interpreted as minimizing the approximation error at a certain sparse level,
\begin{equation}\label{1}
\begin{aligned}
  \hat{\alpha_i}=\text{arg min}\|\bm{D}_i\alpha_i-\bm{y}_i\|_2\\
s.t\quad \|\bm{\alpha}\|_0\leq K_0
\end{aligned}
\end{equation}
where $||*||_{2}$ is the $L_2$ norm and $K_0$ is an upper bound of given sparsity degrees.
Though this minimisation problem is still regarded as a NP-hard problem by computer scientists, and almost impossible to solve, $\alpha_i$ can be approximately derived by greedy pursuit algorithms, such as orthogonal matching pursuit (OMP)~\cite{pati1993orthogonal,tropp2007signal} or subspace pursuit (SP)~\cite{dai2009subspace}.
In this paper, we use the OMP algorithm to solve the above optimization problem.
The OMP algorithm is a refinement of the matching pursuit (MP)~\cite{mallat1993matching}, and its basic idea is described as follows.
A sparse approximation is constructed by selecting an atom matching the signal $\bm{y}_i$ from a dictionary matrix $\bm{D}_i$.
The residual error of signals can be derived, then the atom matching the residual error best is selected contiguously.
The iteration is carried on until the residual error is negligible or the pre-defined iteration number is achieved.
In contrast to the MP algorithm, the OMP algorithm request that all atoms selected at each step of the decomposition should be orthogonal.
The convergence speed of the OMP algorithm is faster than the original matching pursuit when the same accuracy is required.

For our band selection method, after each band image is represented by the corresponding dictionaries, we get a coefficient matrix $\bm{X}\in R^{B\times B}$.
In $\bm{X}$, most of entries would be equal to zero, since the sparsity constraint $K_0$ is added to the equation (8).
The greater the absolute value of coefficients is, the more important the corresponding basis of dictionary is in forming the original hyperspectral image.
For each row of the coefficient matrix $\bm{X}$, we count the number of entries which are not zero.
Then we get a histogram of the corresponding indices which are the band number.
If $h$ denotes the histogram, it is computed by
\begin{equation}
  h=\sum_{i=1}^{B}g\left(\alpha_i)\right/B
\end{equation}
where $g(x)=1$ if $x\neq0$, $g(x)=0$ if $x=0$.
The bands with the first $\tilde{B}$ greater values of histogram series should be selected as the target bands.
The following \textbf{Algorithm 1} gives the overall computing workflow for our band selection method.

\begin{algorithm}[htb]
\caption{Multi-dictionary Sparse Representation Method.}
\label{alg:Framwork}
\begin{algorithmic}[1]
\REQUIRE ~~\\
    Hyperspectral image $\bm{Y}\in R^{W\times H \times B}$,
    number of pixels $N$ ($N\ll B$),
    sparsity level $K_0$,
    target band dimension $\tilde{B}$.
\ENSURE ~~\\
    Reduced-band hyperspectral image $\bm{\tilde{Y}}\in R^{W\times H \times \tilde{B}}$;
\STATE Convert $\bm{Y}$ into  $\bm{Y}=[\bm{y}_1, \bm{y}_2,...,\bm{y}_B]$, $\bm{y}_i \in R^{WH}$;
\STATE Randomly select $N$ pixels in $\bm{Y}$ to form a new sample set $\bm{\hat{Y}}=[\bm{\hat{y}}_1, \bm{\hat{y}}_2,...,\bm{\hat{y}}_B]$, $\bm{\hat{y}}_i \in R^{N}$;
\label{code:fram:extract}
\STATE Construct a group of overcompleted dictionaries,~\emph{i.e.}, $\bm{D}_i=[\bm{\hat{y}}_{1}, \bm{\hat{y}}_{2},..., \bm{\hat{y}}_{i-1}, 0, \bm{\hat{y}}_{i+1},..., \bm{\hat{y}}_{B}]$;
\label{code:fram:trainbase}
\STATE Apply OMP algorithm to Equ.(8) and find the coefficient matrix $\bm{\tilde{X}}\in R^{\tilde{B}\times N}$;
\label{code:fram:add}
\STATE Count nonzero entries in every row of $\bm{\tilde{X}}$ by Equ.(9);
\label{code:fram:classify}
\STATE Sort the number of nonzero entries in descending order.
\label{code:fram:select}
\STATE Select the bands with the first $\tilde{B}$ indices.
\label{code:fram:select}
\STATE Build $\bm{\tilde{Y}}$ with selected bands.
\end{algorithmic}
\end{algorithm}
\section{Experiments}
In this section, we perform extensive experiments to verify the effectiveness of the proposed band selection algorithm.
First, we introduce the evaluation metrics and several other methods that achieve state-of-art performance in hyperspectral image classification.
Then, we present three widely used hyperspectral image datasets for our comparison.
Finally, experimental results and detailed discussions are provided to demonstrate the effectiveness and advantages of the proposed band selection approach.
\subsection{Evaluation Metrics and Compared Methods}
In this paper, the proposed band selection method is mainly used for hyperspectral image classification.
Therefore, the overall classification accuracy (OCA) is the main evaluation metric used in the experimental analysis.
In addition, to compare the performance of each method in eliminating the data redundancy, the correlation coefficient of selected bands and the time of computation are also calculated and compared.

In order to fairly evaluate the performance of the proposed method and other state-of-the-art techniques, unsupervised comparison settings without using labeled information have been made~\cite{du2008similarity,yang2011unsupervised}.
More specifically, we compare several classical band selection methods, \emph{i.e.}, linear prediction (LP) based band selection~\cite{du2008similarity,yang2011unsupervised}, orthogonal subspace projection (OSP) based band selection~\cite{du2008similarity}, and cluster based band selection~\cite{martinez2006clustering}.
In addition, the comparison also includes the traditional dimensionality reduction methods, such as PCA based feature extraction.

\emph{1) LP-based band selection}~\cite{du2008similarity}:
This kind of band selection methods are based on the criterion of band dissimilarity.
Two bands are first selected randomly or based on the dissimilarity among the original band set.
Using the selected bands, the methods estimate the next candidate band by performing linear prediction and least squares solution.
The band that yields the maximum error is considered as the most dissimilar band to the selected band set, and can be selected for the next iteration.
%

\emph{2) OSP-based band selection}~\cite{du2008similarity}:
This kind of methods first compute the orthogonal subspace of selected bands.
Then it compute all other bands' projection in this subspace.
The band that yields the maximum orthogonal component is the most dissimilar band to the selected bands, and can be selected as a new target band.

\emph{3) Cluster-based band selection}~\cite{martinez2006clustering}:
In this kind of methods, every band image is considered a performed object.
First, all bands are clustered using a hierarchical clustering algorithm.
Then, the methods select the most representative band in every cluster based on divergence measures.
These representative bands can be considered as the selected bands.
\subsection{Hyperspectral Datasets}
Three real-world hyperspectral image datasets~\cite{dataset-web} are used for experimental verifications.
All of images in these three datasets are earth observation images taken from airbornes or satellites over some public available hyperspectral scenes.

\emph{1) Salinas-A Scene dataset:} The Salinas-A Scene dataset is a subset of the Salinas Scene dataset, gathered by AVIRIS (Airborne Visible/Infrared Imaging Spectrometer) sensor in 1998, over Salinas Valley in California.
It is characterized by a high spatial resolution (3.7-meter pixels).
The covered area comprises 512 lines by 217 samples. It includes vegetables, bare soils, and vineyard fields.
In Salinas-A, there are $86\times 83$ pixels with 224 bands including 20 bands of water abortion.
Following previous works~\cite{Ma2015Hyperspectral}, we discarded the 20 water absorption bands.
The groundtruth of this dataset contains 6 classes.
Figure 2 shows the Salinas-A grayscale image and its corresponding pseudo color ground truth.
\begin{figure}[H]
\begin{minipage}[t]{0.45\linewidth}
\centering
\includegraphics[width=1\linewidth,height=4.0cm]{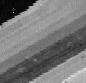}
 \centerline{(a)}\medskip
\end{minipage}
\begin{minipage}[t]{0.45\linewidth}
\centering
\includegraphics[width=1\linewidth,height=4.0cm]{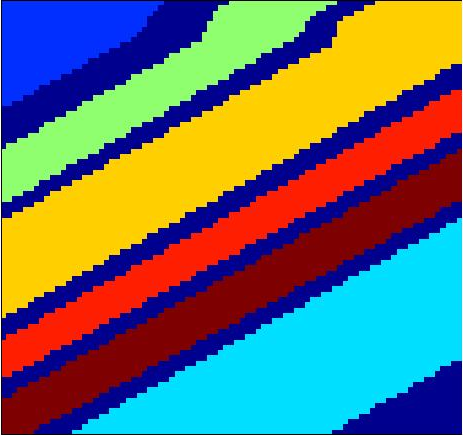}
\centerline{(b)}\medskip
\end{minipage}
\begin{minipage}[c]{1.0\linewidth}
\centering
\includegraphics[width=1\linewidth,height=1.0cm]{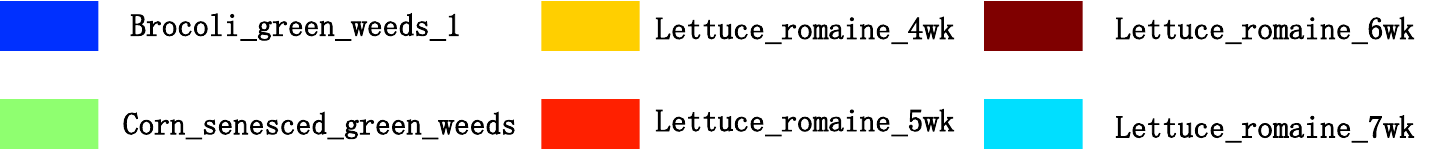}
\end{minipage}
\caption{Sample image in the Salinas-A Scene dataset. (a) Grayscale image. (b) Corresponding pseudo color ground truth.}
\label{fig:res}
\end{figure}
\begin{figure}[H]
\begin{minipage}[t]{0.45\linewidth}
\centering
\includegraphics[width=1\linewidth,height=7.0cm]{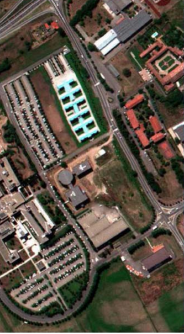}
 \centerline{(a)}\medskip
\end{minipage}
\begin{minipage}[t]{0.45\linewidth}
\centering
\includegraphics[width=1\linewidth,height=7.0cm]{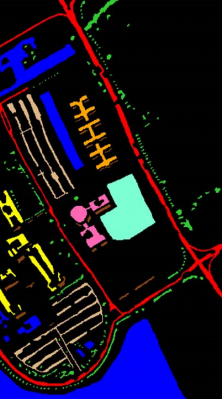}
\centerline{(b)}\medskip
\end{minipage}
\begin{minipage}[c]{1.0\linewidth}
\centering
\includegraphics[width=1\linewidth,height=1.0cm]{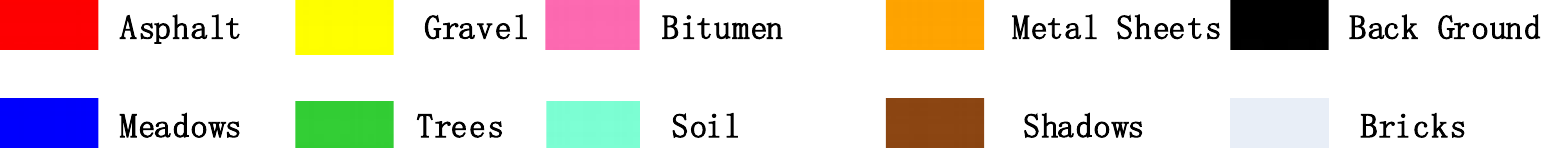}
\end{minipage}
\caption{Sample image in the Pavia University Scene dataset. (a) True color image. (b) Corresponding pseudo color ground truth.}
\label{fig:res}
\end{figure}

\emph{2) Pavia University Scene dataset:} The Pavia University Scene dataset was acquired by the ROSIS (Reflective Optics System Imaging Spectrometer) sensor over the Engineering School, Pavia University, northern Italy.
The number of spectral bands is 103, and the number of pixels is 610$\times$340.
The geometric resolution of this scene is 1.3 meters.
Pavia University Scene dataset employed in this paper is the one with groundtruth land cover map, which includes 9 classes.
Figure 3 shows the image of this dataset and its corresponding pseudo color ground truth.

\emph{3) Indian Pines Scene dataset:} This dataset was also gathered by AVIRIS sensor in 1992, over the Indian Pines test site in north-western Indiana.
It consists of 145$\times$145 pixels and 224 spectral reflectance bands in the wavelength range of 0.4 - 2.5 $\mu$m.
The Indian Pines Scene dataset contains two-thirds agriculture, and one-third forest or other natural perennial vegetation.
There are two major dual lane highways, a rail line, as well as some low density housing, other built structures, and smaller roads.
This scene is a subset of a larger scene.
The ground truth available is designated into 16 classes and is not all mutually exclusive.
In the experiments, 12 classes are selected.
Following previous works~\cite{chen2011hyperspectral,Gualtieri1998Support}, we also reduce the number of bands to 200 by removing bands covering the region of water absorption.
Figure 4 shows this scene's grayscale image and its ground truth.
\begin{figure}[H]
\begin{minipage}[t]{0.45\linewidth}
\centering
\includegraphics[width=1\linewidth,height=5.0cm]{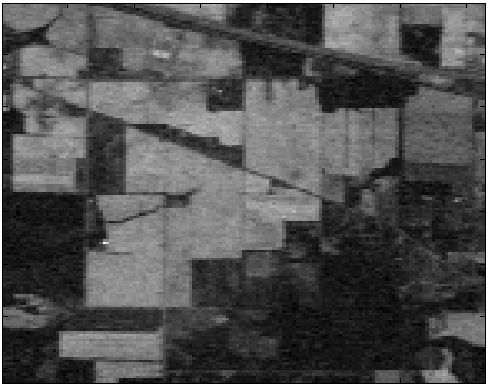}
 \centerline{(a)}\medskip
\end{minipage}
\begin{minipage}[t]{0.45\linewidth}
\centering
\includegraphics[width=1\linewidth,height=5.0cm]{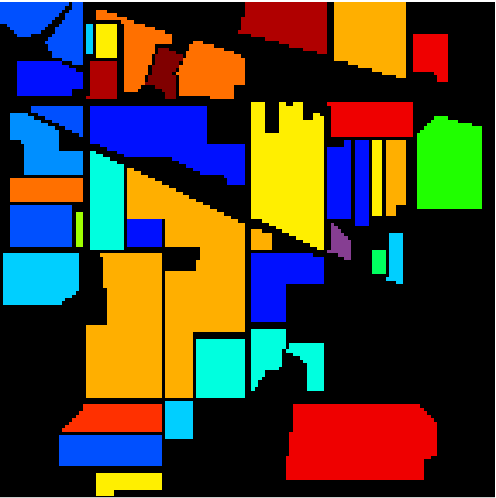}
\centerline{(b)}\medskip
\end{minipage}
\begin{minipage}[c]{1.0\linewidth}
\centering
\includegraphics[width=1\linewidth,height=3.0cm]{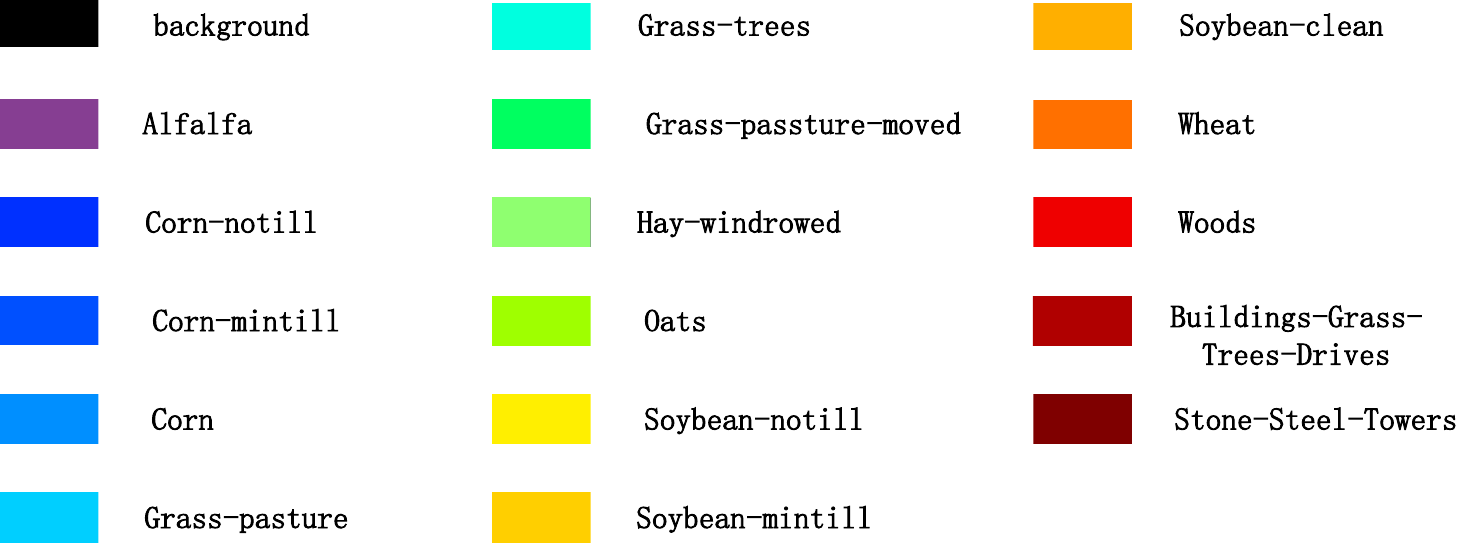}
\end{minipage}
\caption{Sample image in the Indian Pines Scene dataset. (a) Grayscale image. (b) Corresponding pseudo color ground truth.}
\label{fig:res}
\end{figure}
\subsection{Classifier Description}
In order to evaluate the classification performance of each band selection method, we resort to the multi-class classifiers with the labeled images and selected bands.
Following previous works~\cite{chang1999joint,Yuan2014Hyperspectral}, both K-nearest neighborhood (KNN) and support vector machine (SVM) are used to compare the classification performance of the selected image bands obtained by various band selected methods.
We use the KNN and SVM methods proposed in~\cite{cover1967nearest} and~\cite{Chang2011LIBSVM}, respectively.
The classification process of hyperspectral images are introduced briefly as follows.

\emph{1) K-Nearest Neighborhood (KNN)}~\cite{cover1967nearest}:
For a testing sample (a pixel in this paper), the \emph{K} nearest neighbor classification algorithm is to find \emph{K} training samples which are closest to the testing sample.
The category of the testing sample is determined by a majority-voting scheme, using the categories of the \emph{K} training samples.
When sufficient training samples are available, the KNN algorithm can achieve higher performance than vanilla supervised classifiers.

\emph{2) Support Vector Machine (SVM)}~\cite{Chang2011LIBSVM,Gualtieri1998Support}:
SVM is a popular supervised learning model that can analyze data, identify patterns, and be used for classification and regression analysis.
In our experiment, a set of one-vs-all SVMs based on the polynomial and RBF kernels have been tested.
One of the advantages of SVMs is that it is not limited by the number of samples and the dimensionality of samples.

\begin{figure*}
\begin{center}
\begin{tabular}{@{}c@{}c@{}c@{}c}
\includegraphics[width=0.32\linewidth,height=4.545cm]{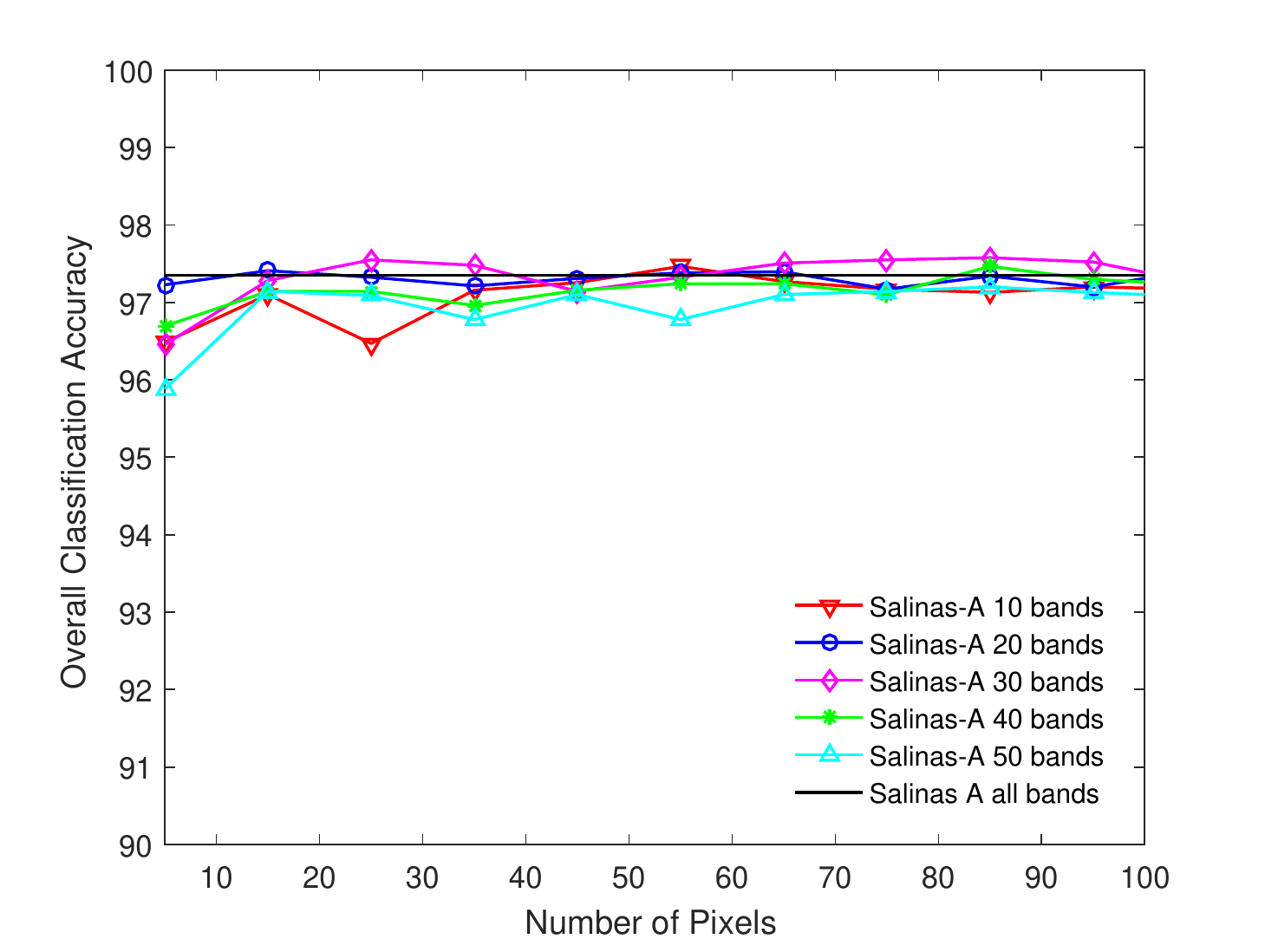} \ &
\includegraphics[width=0.32\linewidth,height=4.545cm]{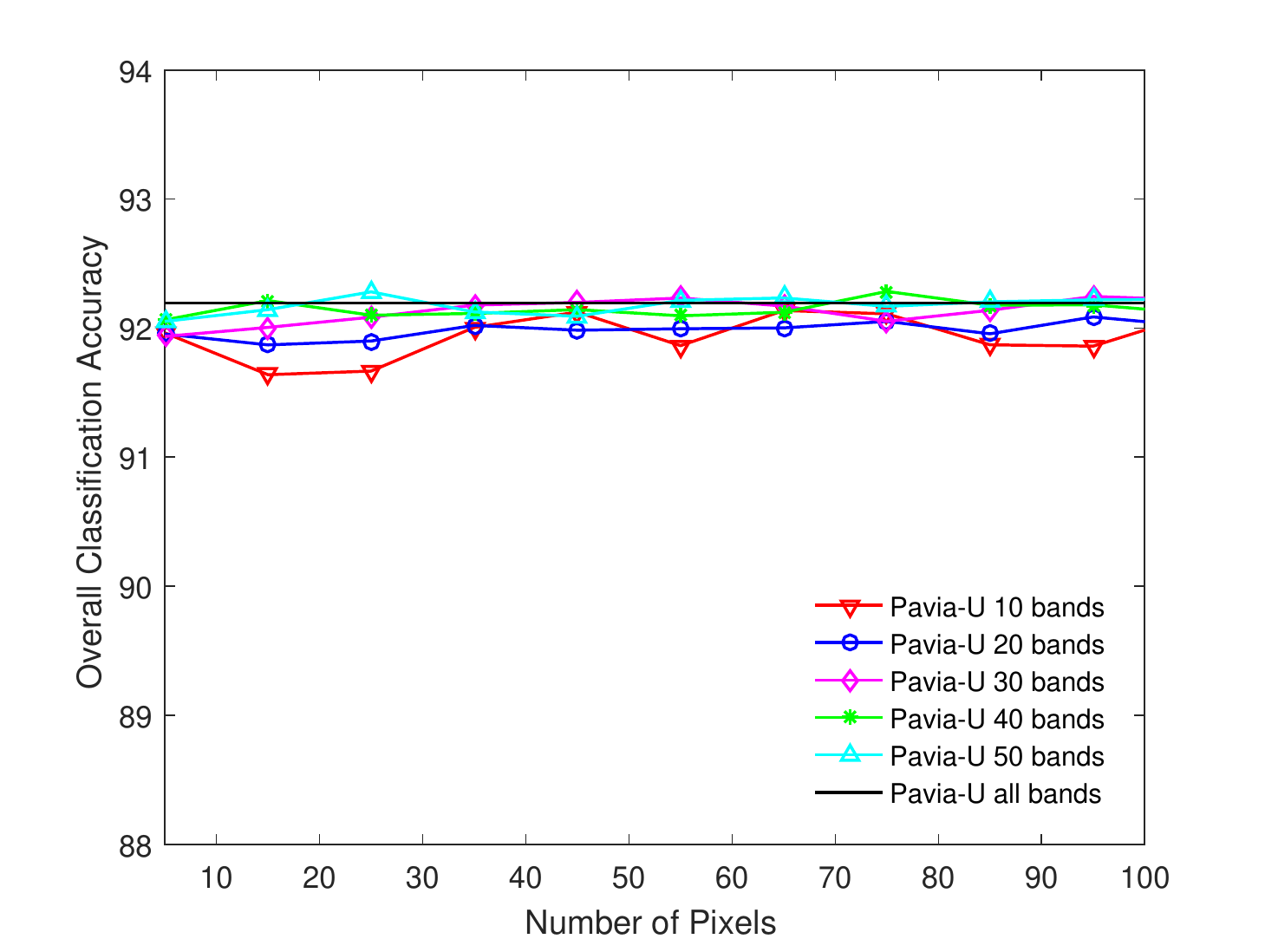} \ &
\includegraphics[width=0.32\linewidth,height=4.545cm]{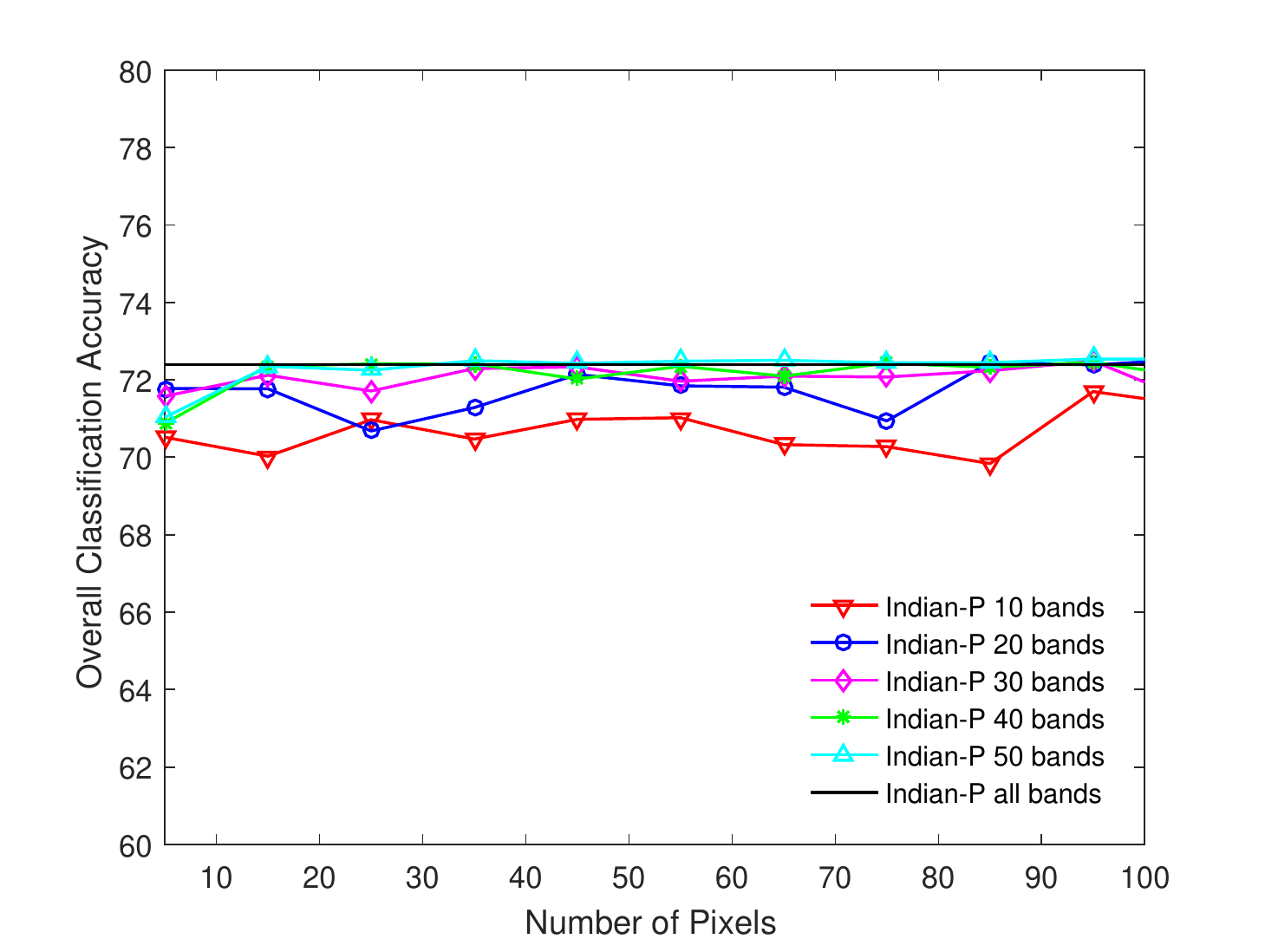} \ \\
 {\small(a) Salinas-A Scene dataset} & {\small(b) Pavia University Scene dataset} & {\small(c) Indian Pines Scene dataset}\\
\end{tabular}
\caption{Illustration of the effect on the overall classification accuracy (OCA) with different numbers of sampling pixels. The classifier is KNN, and $K$ is set to 6, 9 and 12 respectively. Sparse level $K_0$ is set to 6.
\label{fig:PR-curve}}
\end{center}
\end{figure*}
\subsection{Sample Number Analysis}
In the proposed band selection method, several dictionaries should be created during the optimization process, meanwhile these dictionaries must be overcompleted.
Since the number of pixels in hyperspectral images is often greater than the number of bands, if all the pixels are involved in the optimization, the dictionaries are not overcompleted definitely.
In order to ensure the completeness of the dictionaries, we select a part of the pixels in the entire image as processing objects, omitting the rest before the optimization.
The 2-6 steps of Algorithm 1 show the optimization process.
Therefore, what we concern in this subsection is the effect of the sample number on the band selection results, reflected on the OCA metric.

To analyze the effect, we perform ablation studies on three datasets.
Specifically, we randomly pick out 20 samples from each class to form the training set.
Using our proposed method, we select 10, 20, 30, 40 and 50 bands from all bands for the experiments.
Thus, each dataset has five band selection results.
To investigate the effect of different numbers of samples on the classification accuracy, we change the sample number from 5 to 100.
Figure 5 illustrates the classification results of the KNN classifiers (For Sailias-A, $K=6$; For Pavia-U, $K=9$; For Indian-P, $K=12$.) using the OCA metric.
From Figure 5, it can be seen that the classification accuracy almost remains unchanged with the increase of sample number, and only minor fluctuations (the value is not greater than 2\%) appear at several specific locations.
On the Salinas-A dataset, the changes in the number of pixels have little effect on the classification results.
On the other two datasets, the overall trend of the result curve is still horizontal.
In addition, on the Salina-A and Pavia-U datasets, the classification accuracy of the band selection is very close to the classification accuracy of the whole bands during changing the number of sampling points.
The variance of classification accuracy on the Indian-P dataset is larger than the other two datasets.
The main reason may be that there are more classes, and samples are more irregularly distributed in the Indian-Pines dataset.
\begin{table}[!htb]
\centering
\caption{Overall classification accuracy (OCA) with the specific number of pixels on three datasets.}
\begin{tabular}{|c|c|c|c|c|c|c|c|c|}
\hline
Number of pixels & 15& 25 & 35 & 45\\
\hline Salinas-A(\%) & 97.1 & 96.47 & 97.16 & 97.25\\
\hline Pavia-U(\%) & 92.15&92.04 & 92.20 & 92.08\\
\hline Indian-P(\%) & 70.03&70.96 & 70.46 & 70.98\\
\hline Number of pixels &55&65&75&85\\
\hline Salinas-A(\%)  & 97.46& 97.27&97.17&97.13\\
\hline Pavia-U(\%)  &92.23& 92.56& 92.07&93.28\\
\hline Indian-P(\%) & 71.02&70.32 &70.545 & 70.27\\
\hline
\end{tabular}
\end{table}

Table I provides more accurate classification results with specific sampling settings when 10 bands are selected.
From the results in Figure 5 and Table I, we can draw the conclusion that the number of sampling points has little effect on the results of band selection.
Thus, a small number of sampling points can be selected for reducing the computation.
In the following experiments, we empirically set the number of pixels to be 50 ($N=50$) for classification evaluation, without further parameter tuning.
\subsection{Selected Band Analysis}
In Algorithm 1, the coefficient matrix $\bm{\tilde{X}}$ is the key support of band selection.
To figure out which kind of bands our method has selected, we draw bar diagrams which illustrate the ratio of the count of non-zero entries to the count of all entries in one column of coefficient matrix $\bm{\tilde{X}}$.
The calculation method is given in Algorithm 1 and Equation (9).
This ratio represents the contribution of one band in the hyperspectral image to the whole image, or the proportion of bands associated with this band in the image.
More specifically, in the coefficient matrix $\bm{\tilde{X}}$, the higher ratio of non-zero entries in the $i$-th column means the larger contribution of the $i$-th band to the whole image.
Figure 6 illustrates the bar diagrams in which 50 sampling pixels are selected randomly and the sparsity level $K_0$ is 6 in solving the sparse optimization problem.
In Figure 6, we can see that the contribution of each band to the original hyperspectral image is very different.
In the Salinas-A dataset, the weights of two specific bands are close to one, and weights of most bands are below 0.5.
In the Pavia-U dataset, the weight of only one band is close to 0.7, and weights of almost all bands are below 0.45.
In the Indian-P dataset, most weights of the bands are zeros.
Table 2 lists the top 5 weights of all bands in the three datasets.
The weight distribution is not uniform over the bands.
These experimental results further indicate that there are many redundancies in the hyperspectral bands and only a few bands are useful.
Therefore, it is reasonable to select appropriate bands for specific applications.
\begin{table}[!htb]
\centering
\caption{The TOP 5 Selected bands and their weights}
\label{my-label}
\begin{tabular}{|l|l|l|l|l|l|l|}
\hline
\multirow{2}{*}{Salinas-A}
& Band Index & 32 &45  &3  &153  &39  \\ \cline{2-7}
& Weight &0.985  &0.980  & 0.525 & 0.485 &0.446 \\ \hline
\multirow{2}{*}{Pavia-U}
& Band Index &91  &59  &3  &1  &87  \\ \cline{2-7}
& Weight & 0.670 &0.456 &0.408  &0.359 &0.330  \\ \hline
\multirow{2}{*}{Indian-P}
& Band Index &42  &29  &35 &6  &1  \\ \cline{2-7}
& Weight & 0.995 &0.905  &0.695  &0.670  &0.590  \\ \hline
\end{tabular}
\end{table}
\begin{figure*}
\begin{center}
\begin{tabular}{@{}c@{}c@{}c@{}c}
\includegraphics[width=0.32\linewidth,height=4.545cm]{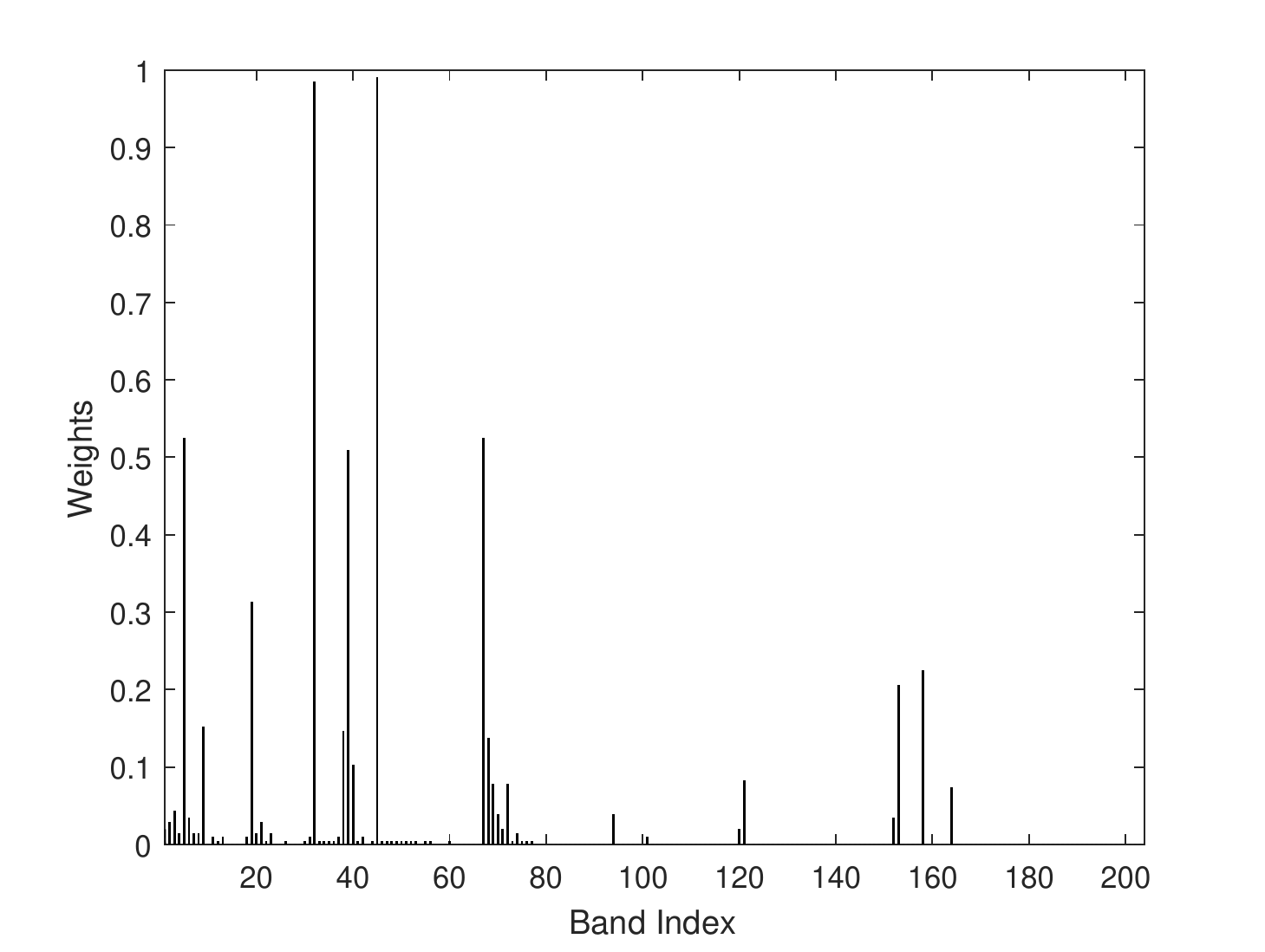} \ &
\includegraphics[width=0.32\linewidth,height=4.545cm]{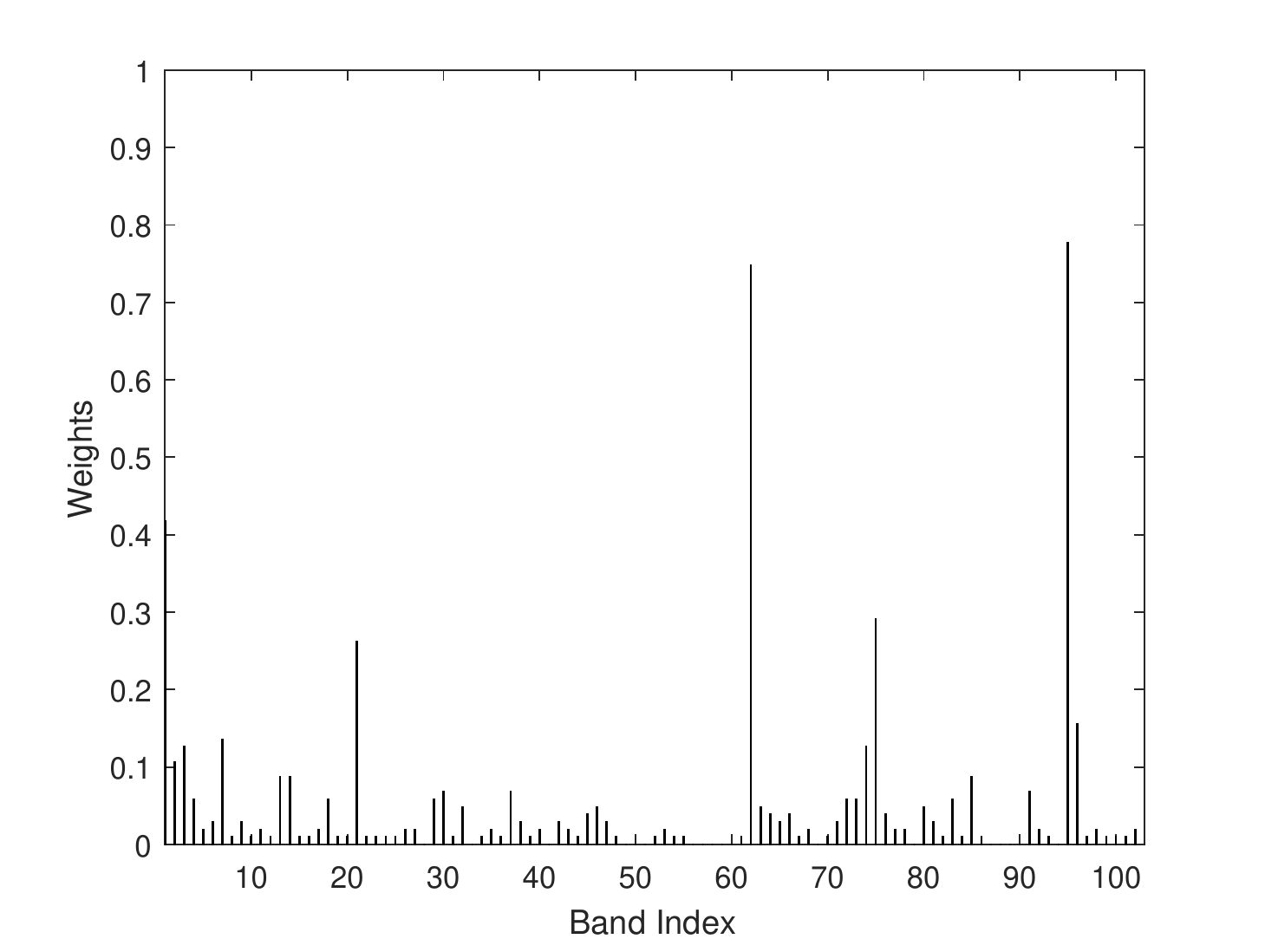} \ &
\includegraphics[width=0.32\linewidth,height=4.545cm]{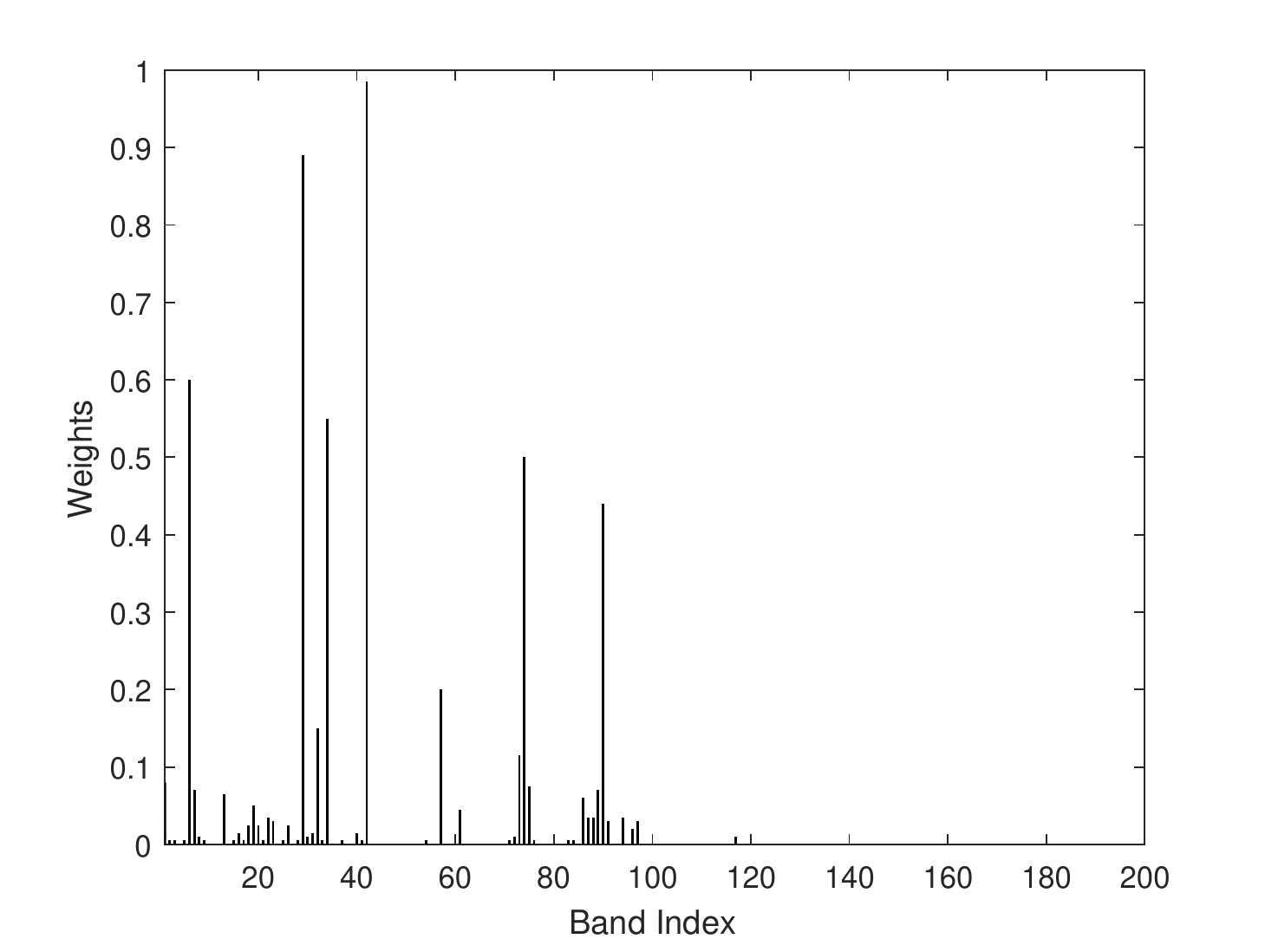} \ \\
 {\small(a) Salinas-A Scene dataset} & {\small(b) Pavia University Scene dataset} & {\small(c) Indian Pines Scene dataset}\\
\end{tabular}
\caption{The histograms of coefficients on the three datasets. The distribution of the band contribution is nonuniform. Most weights of bands are very low.
\label{fig:PR-curve}}
\end{center}
\end{figure*}
\begin{figure*}
\begin{center}
\begin{tabular}{@{}c@{}c@{}c@{}c}
\includegraphics[width=0.32\linewidth,height=4.545cm]{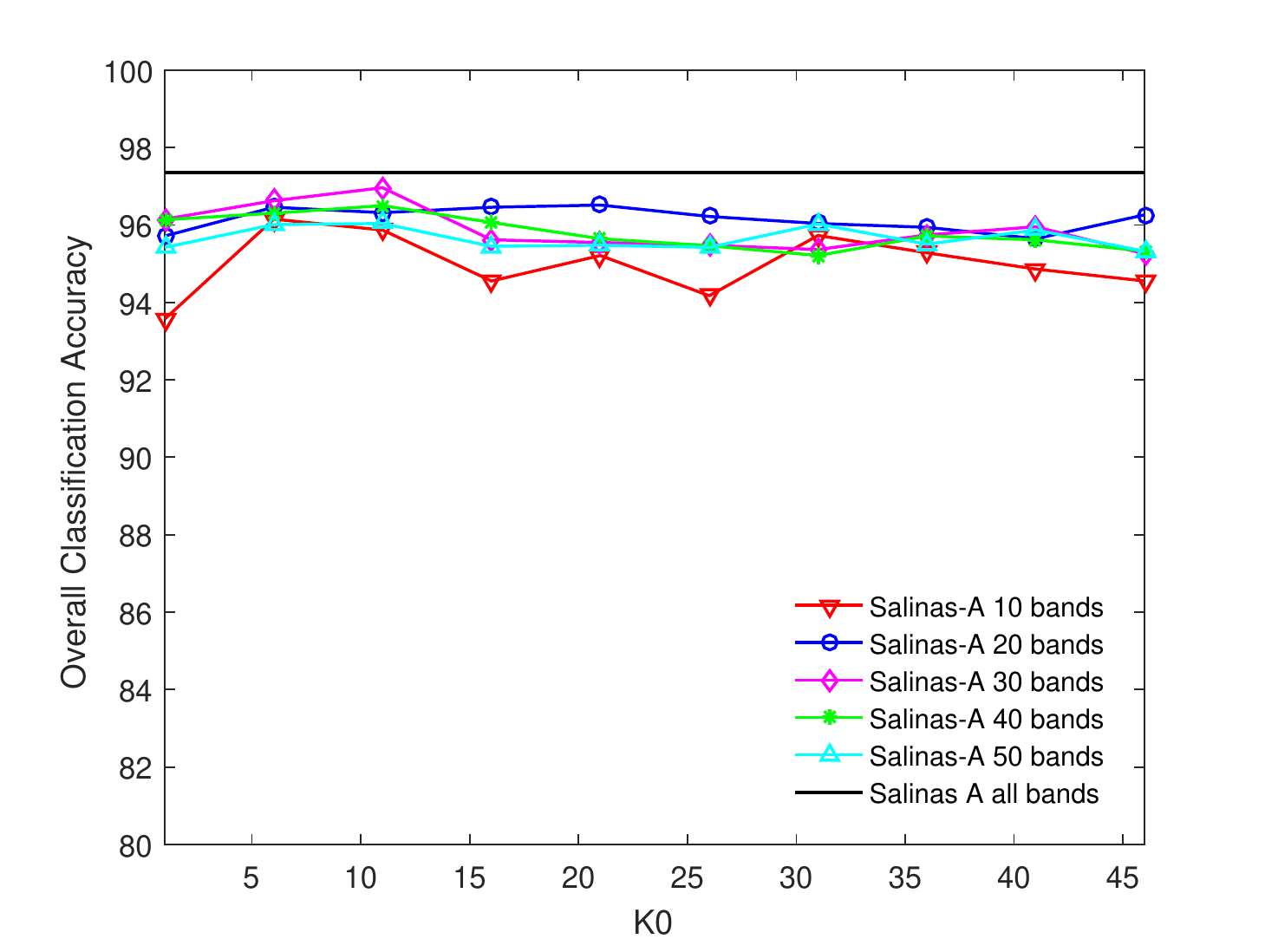} \ &
\includegraphics[width=0.32\linewidth,height=4.545cm]{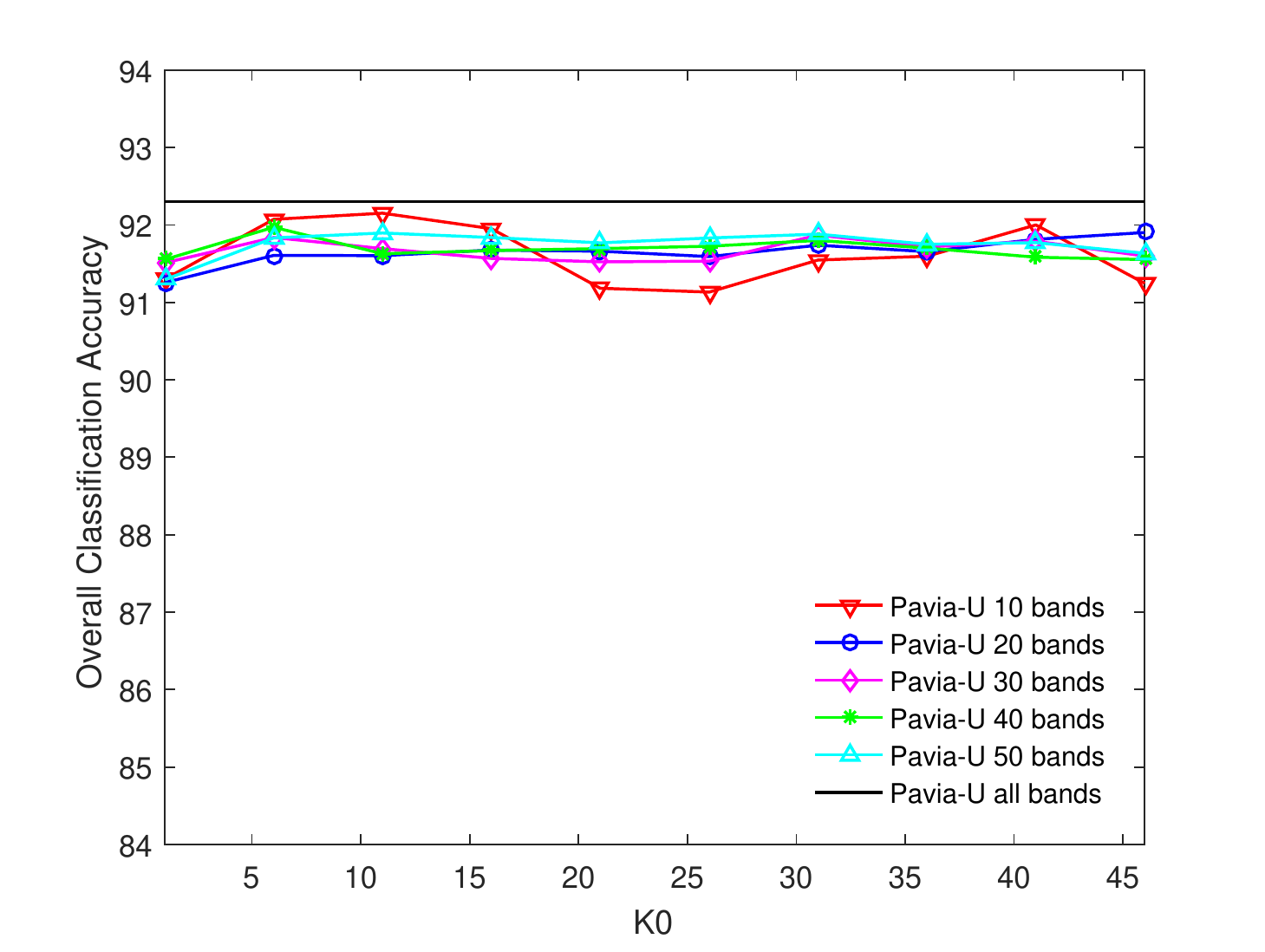} \ &
\includegraphics[width=0.32\linewidth,height=4.545cm]{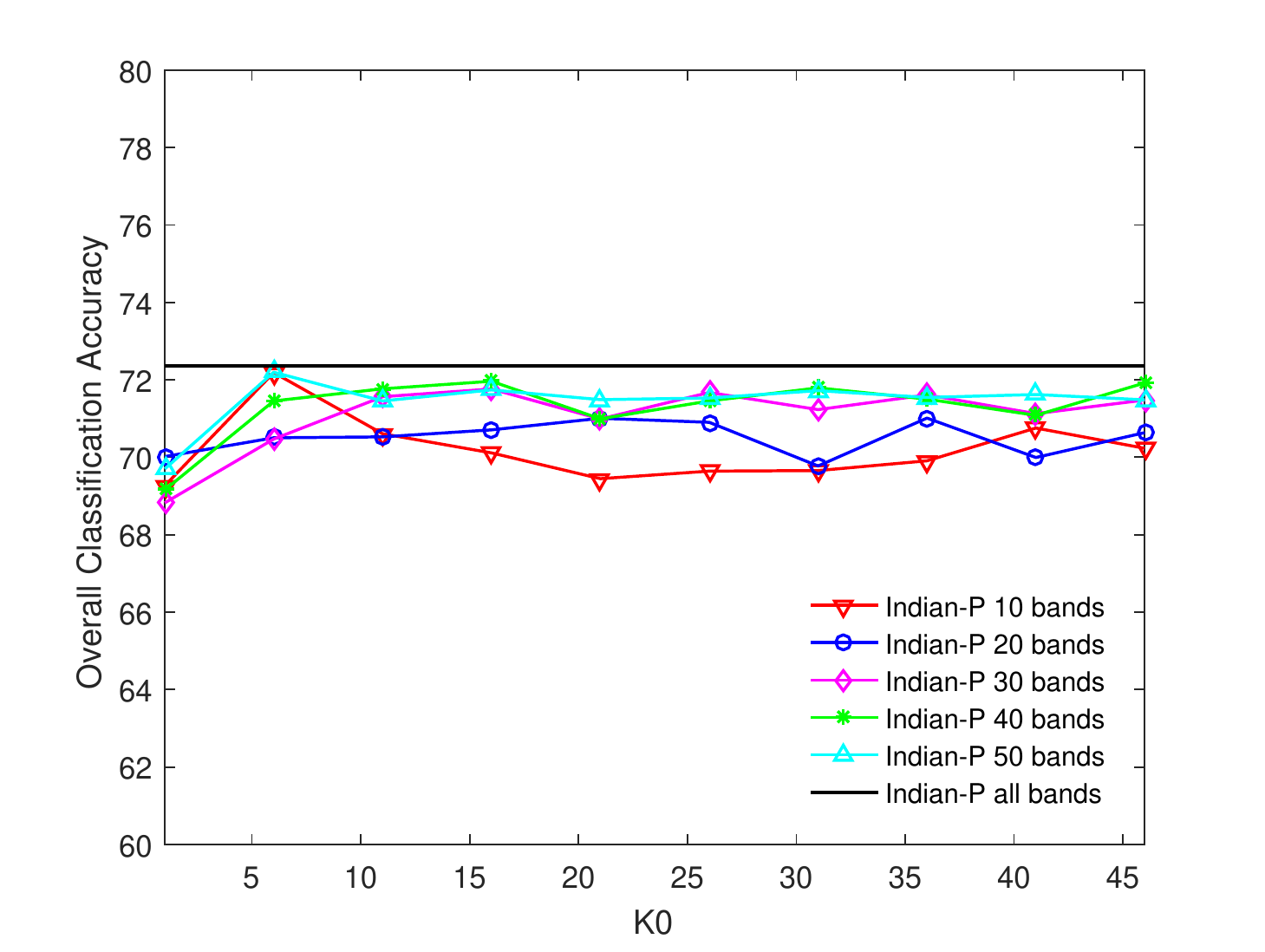} \ \\
 {\small(a) Salinas-A Scene dataset} & {\small(b) Pavia University Scene dataset} & {\small(c) Indian Pines Scene dataset}\\
\end{tabular}
\caption{The overall classification accuracy with different sparsity levels. We use the KNN classifier on the three datasets.
\label{fig:PR-curve}}
\end{center}
\end{figure*}
\begin{figure*}
\begin{center}
\begin{tabular}{@{}c@{}c@{}c@{}c}
\includegraphics[width=0.32\linewidth,height=4.545cm]{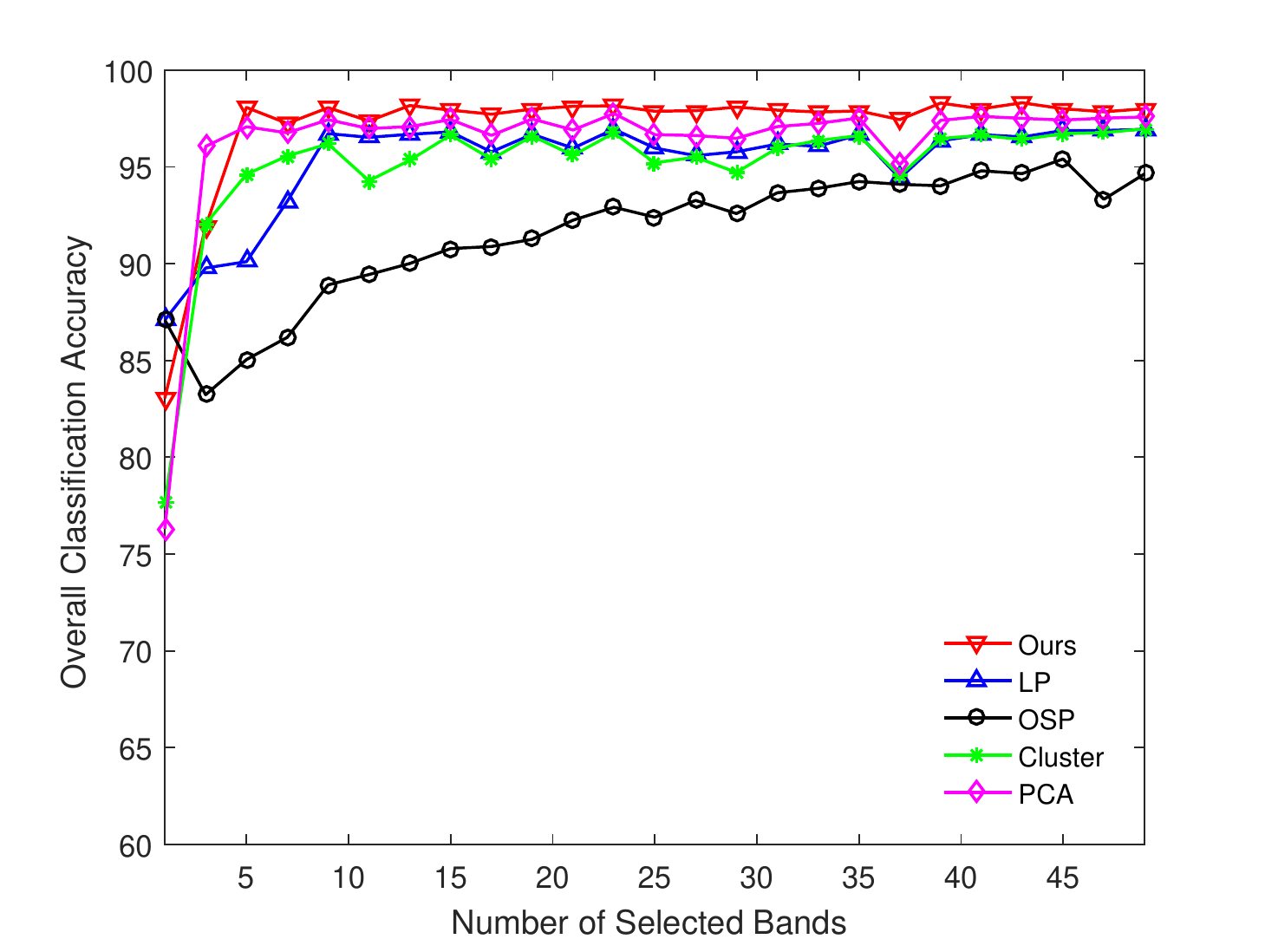} \ &
\includegraphics[width=0.32\linewidth,height=4.545cm]{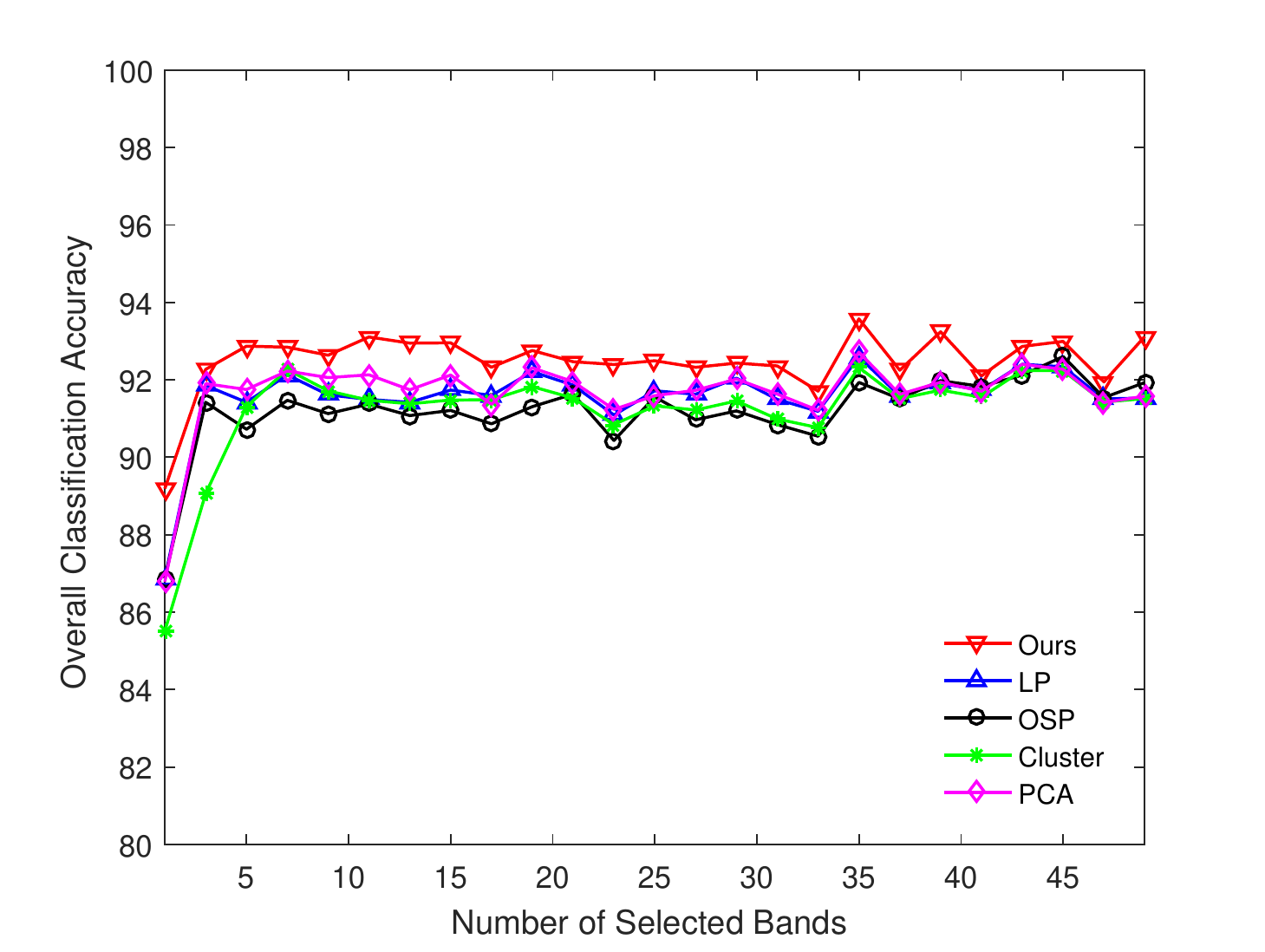} \ &
\includegraphics[width=0.32\linewidth,height=4.545cm]{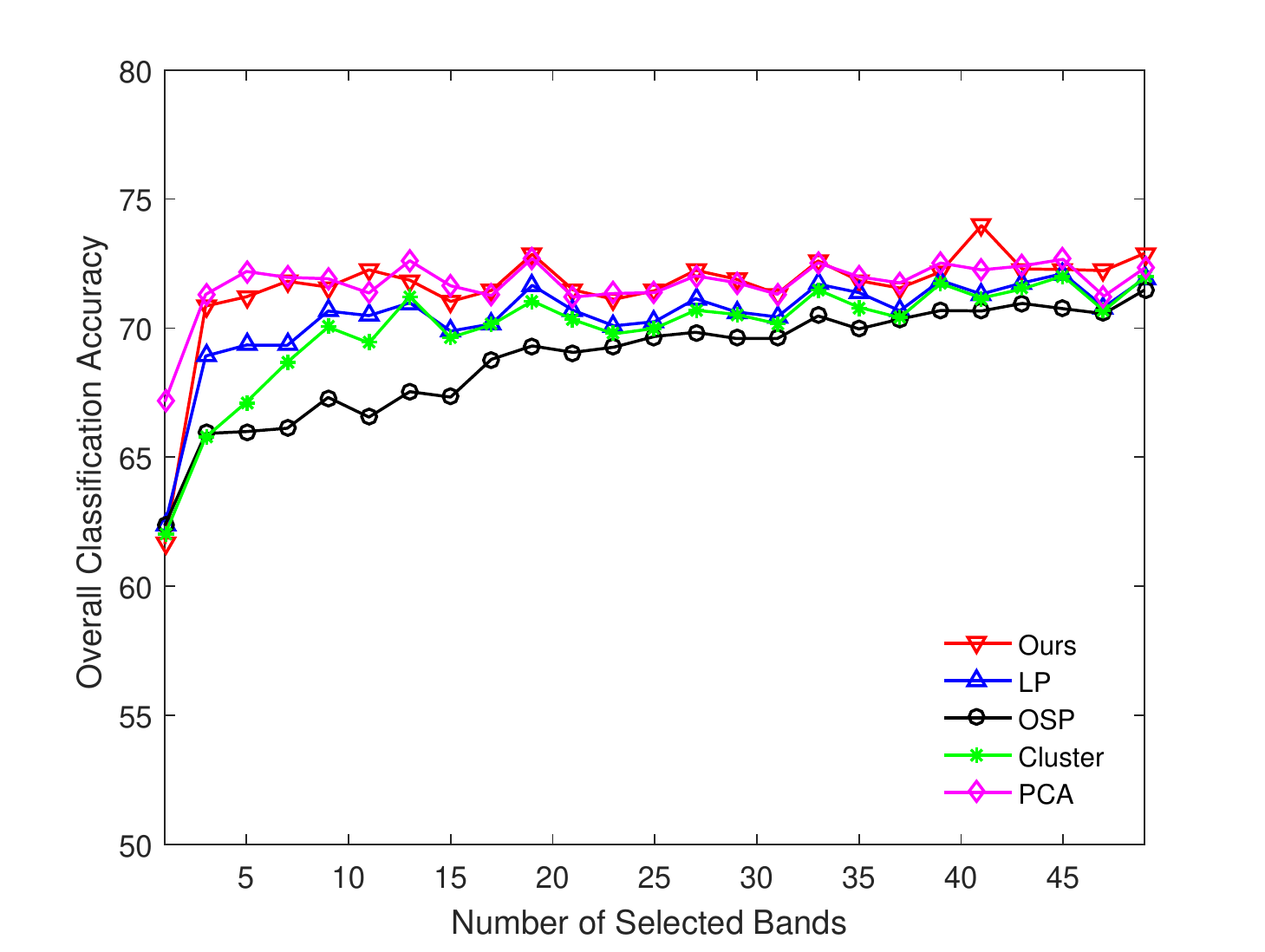} \ \\
 {\small(a) Salinas-A Scene dataset} & {\small(b) Pavia University Scene dataset} & {\small(c) Indian Pines Scene dataset}\\
\end{tabular}
\caption{The classification results of different band selection methods using the KNN classifier on the three datasets.
\label{fig:PR-curve}}
\end{center}
\end{figure*}
\begin{figure*}
\begin{center}
\begin{tabular}{@{}c@{}c@{}c@{}c}
\includegraphics[width=0.32\linewidth,height=4.545cm]{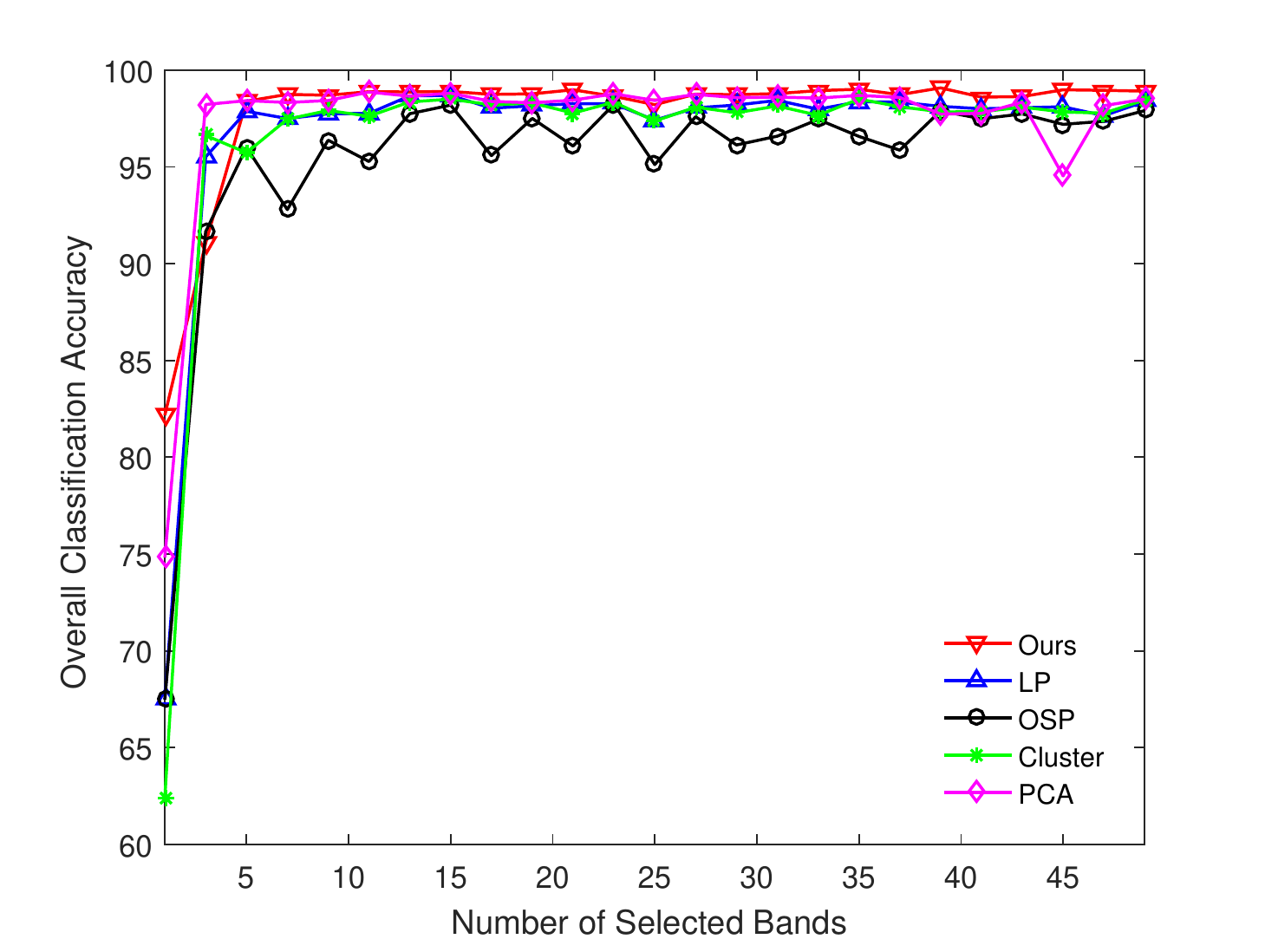} \ &
\includegraphics[width=0.32\linewidth,height=4.545cm]{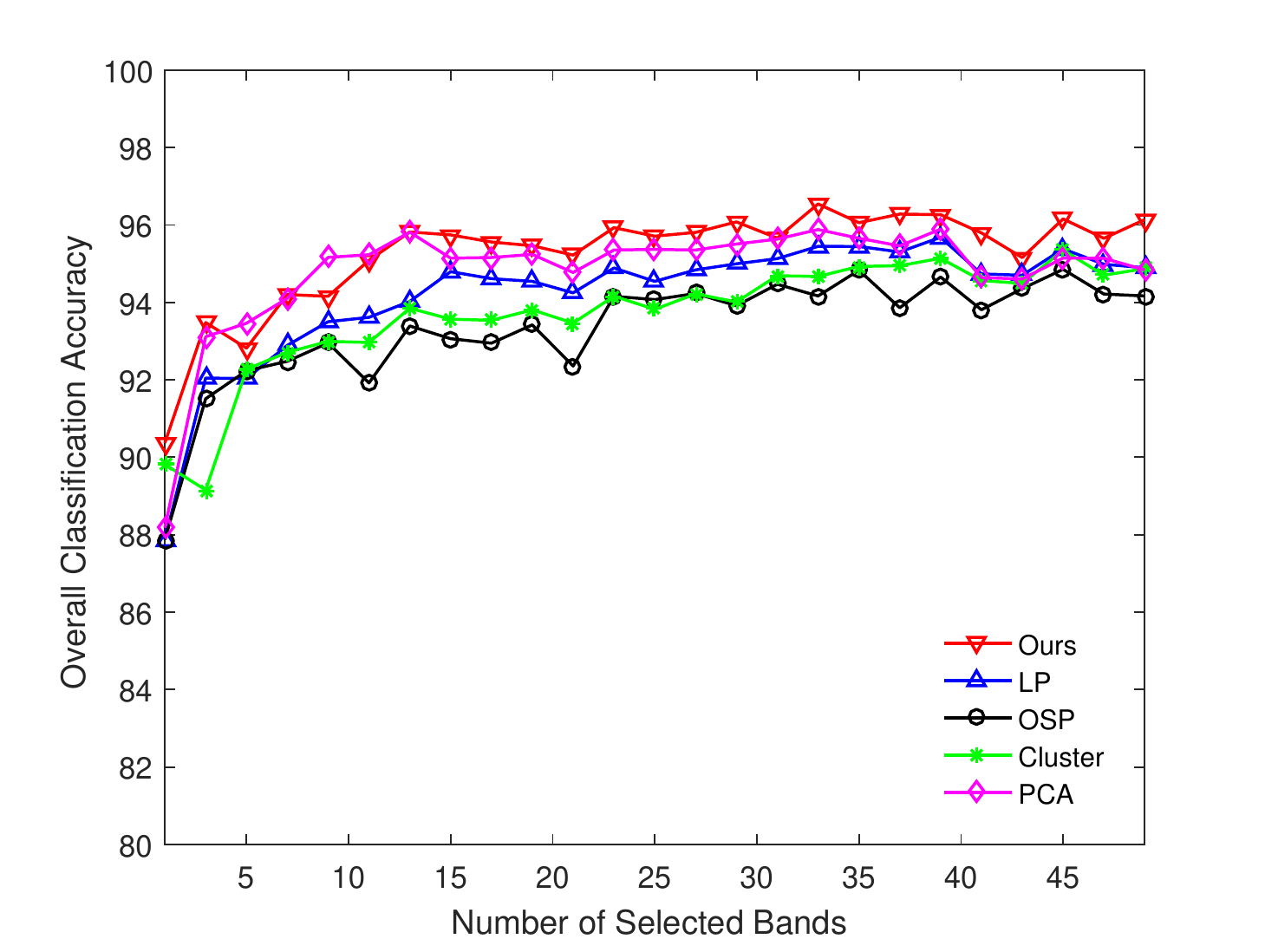} \ &
\includegraphics[width=0.32\linewidth,height=4.545cm]{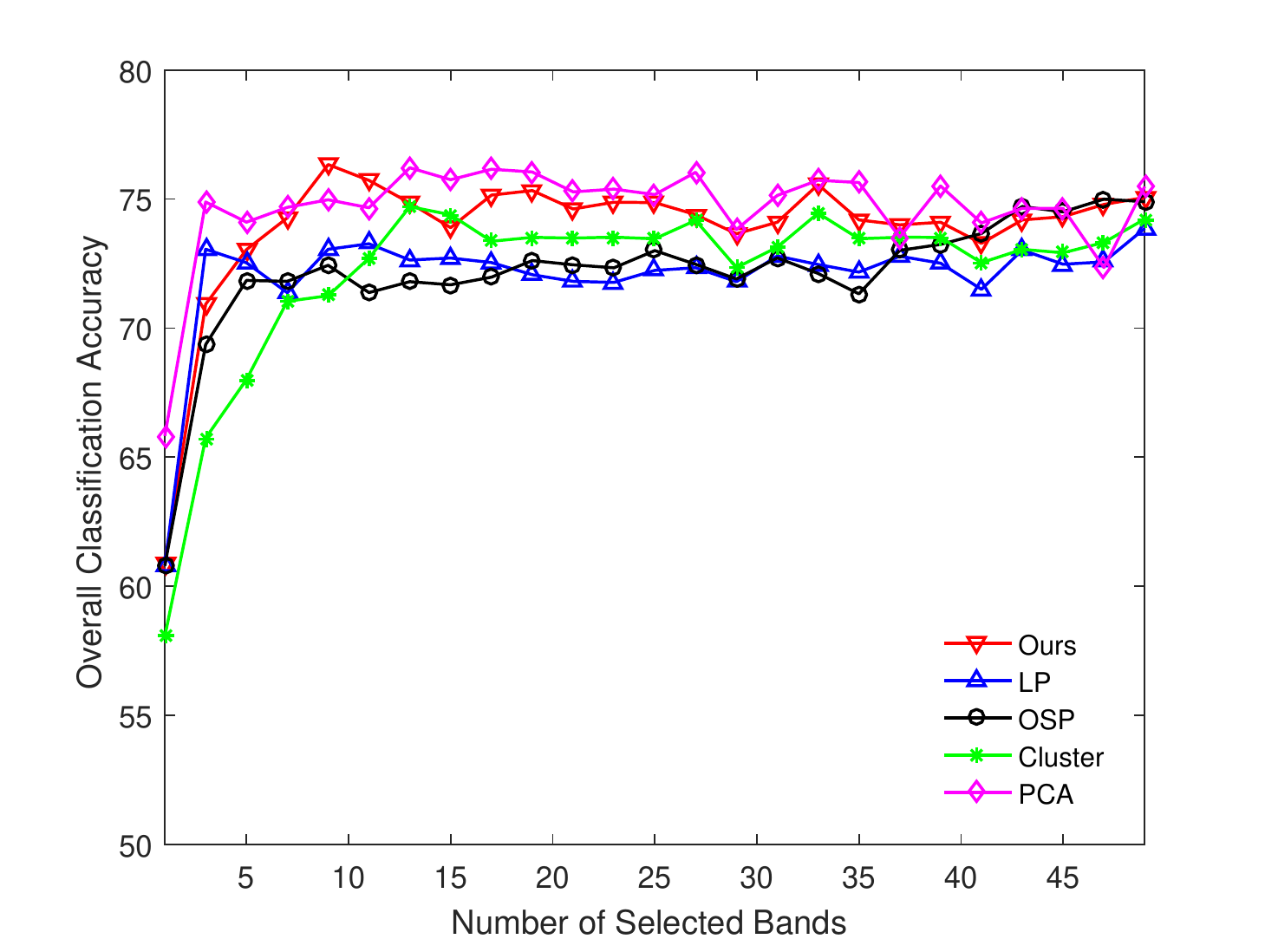} \ \\
 {\small(a) Salinas-A Scene dataset} & {\small(b) Pavia University Scene dataset} & {\small(c) Indian Pines Scene dataset}\\
\end{tabular}
\caption{The classification results of different band selection methods using the SVM classifier on the three datasets.
\label{fig:PR-curve}}
\end{center}
\end{figure*}
\subsection{Sparse Level Analysis}
In solving the coefficient matrix $\bm{\tilde{X}}$, the sparsity level $K_0$ should be set appropriately.
To investigate its impact, we evaluate a set of choices of $K_0$ on the three datasets, where we use the KNN classifier and $K$ is set to be $6, 9, 12$, respectively.
Figure 7 shows the classification accuracy curves when the sparsity level $K_0$ changes from 1 to 50 on the three datasets.
There are several fundamental observations in Figure 7:
1) for the Salinas-A dataset, when $K_0$ is less than 3, the classification accuracy is relatively low.
When $K_0$ is greater than 3, the classification performance first increases, then drops down slightly.
The classification performance of the model with 10 bands fluctuates more drastically than other models.
This indicates that the model with very few bands is sensitive to the sparse level $K_0$.
The same trend can be observed in other two datasets.
2) for the Pavia-U dataset, the classification performance with different sparse levels and selected bands remains almost unchanged (no more than 2\% fluctuation), even they vary locally.
This result shows that our method is very robust in this dataset under different sparse levels.
3) for the Indian-P dataset, the models with more bands are better in the classification performance.
The models are more fluctuated on this dataset under different $K_0$.
Based on above facts, we set $K_0=6$ in the following experiments.
\subsection{Comparison with Other Methods}
In order to fairly evaluate the performance of different band selection algorithms, KNN and SVM classifiers are adopted to classify the hyperspectral image after the band selection.
The value of $K$ in KNN is set to 6, 9 and 12 for the Salinas-A, Pavia-U and Indian-P datasets, respectively.
The parameters of the SVM classifier is obtained by the $K$-fold cross validation.
The compared methods have been described in section III. A.
{\flushleft\textbf{Overall Classification Performance}}
For the classification performance evaluation, 20 samples are selected randomly in each classes.
For our method, we set $K_0=6$.
In order to reduce the randomness, we perform all the compared methods with 10 trials and use the averaged results.
We report the results with different selected band number $n$, ranging from 1 to 50.
The Figure 8 and Figure 9 show the classification performance of the proposed method and compared algorithms on the three datasets.
From the comparison results, we can see that our proposed method consistently outperforms other methods by a large margin, especially on the Salinas-A and Pavia-U datasets.
More specifically, the OSP method achieves the lowest recognition rate.
The LP and cluster based methods have similar classification results, which are inferior to our method .
Even though the PCA method is superior to other methods, it is not better than our proposed method.

To further verify the advantages of our proposed method, we also calculate the Kappa coefficient, which measures the agreement between two raters who each classify N items into C mutually exclusive categories.
The results with the KNN and SVM classifiers are shown in Figure 10 and Figure 11, respectively.
From the results, we can see that our method also achieve better performance with the Kappa coefficient measure.
In summary, the proposed method is better than most of compared methods in terms of both classification accuracy and Kappa coefficient.
Our method is inferior to the PCA method in several cases on the Indian-P dataset, however, the performance is still very comparative.
{\flushleft\textbf{Band Correlation Analysis}}
To measure the separation of selected bands, we calculate the correlation coefficient of bands selected by different methods.
Generally, the lower the correlation among spectral bands is, the better the subset of selected bands is.
Figure 12 illustrates the average correlation coefficient with the compared methods other than PCA.
On the three data subsets, our method ranks first or second in this metric, which is in agreement with previous results shown in Figure 8-11.
This further confirms that our method is able to select the most informative and separative bands for the real-world applications.
\begin{figure*}
\begin{center}
\begin{tabular}{@{}c@{}c@{}c@{}c}
\includegraphics[width=0.32\linewidth,height=4.545cm]{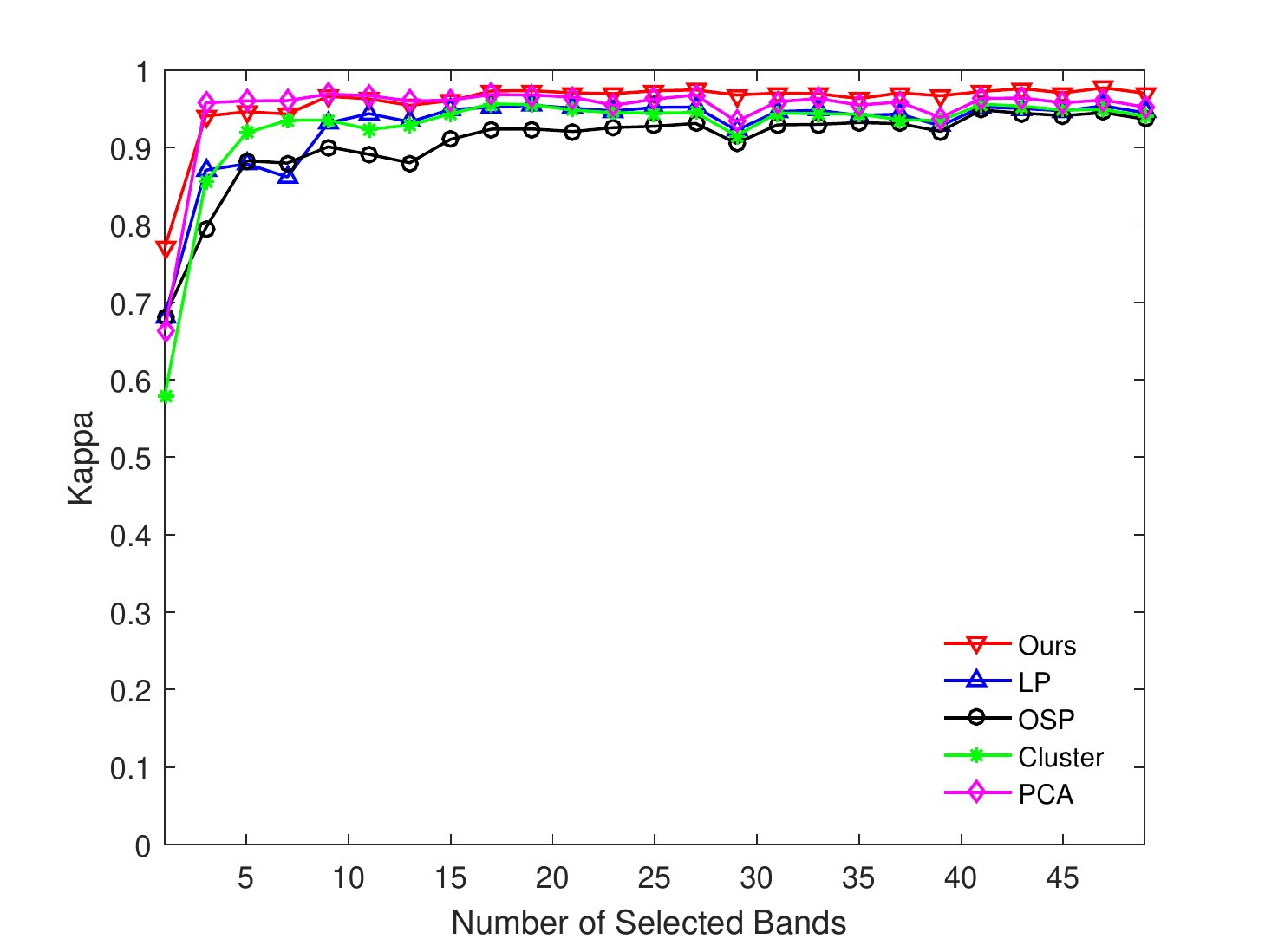} \ &
\includegraphics[width=0.32\linewidth,height=4.545cm]{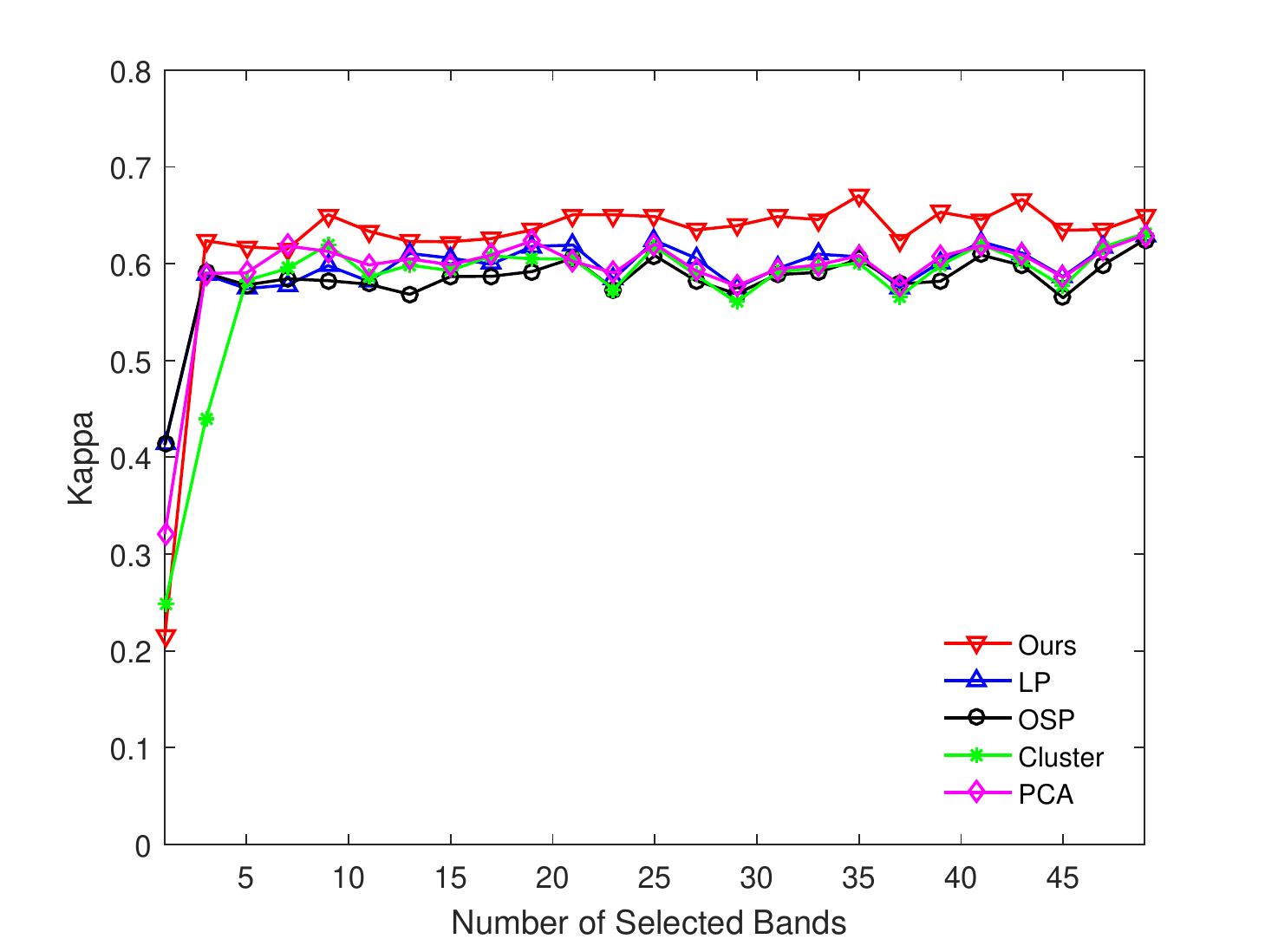} \ &
\includegraphics[width=0.32\linewidth,height=4.545cm]{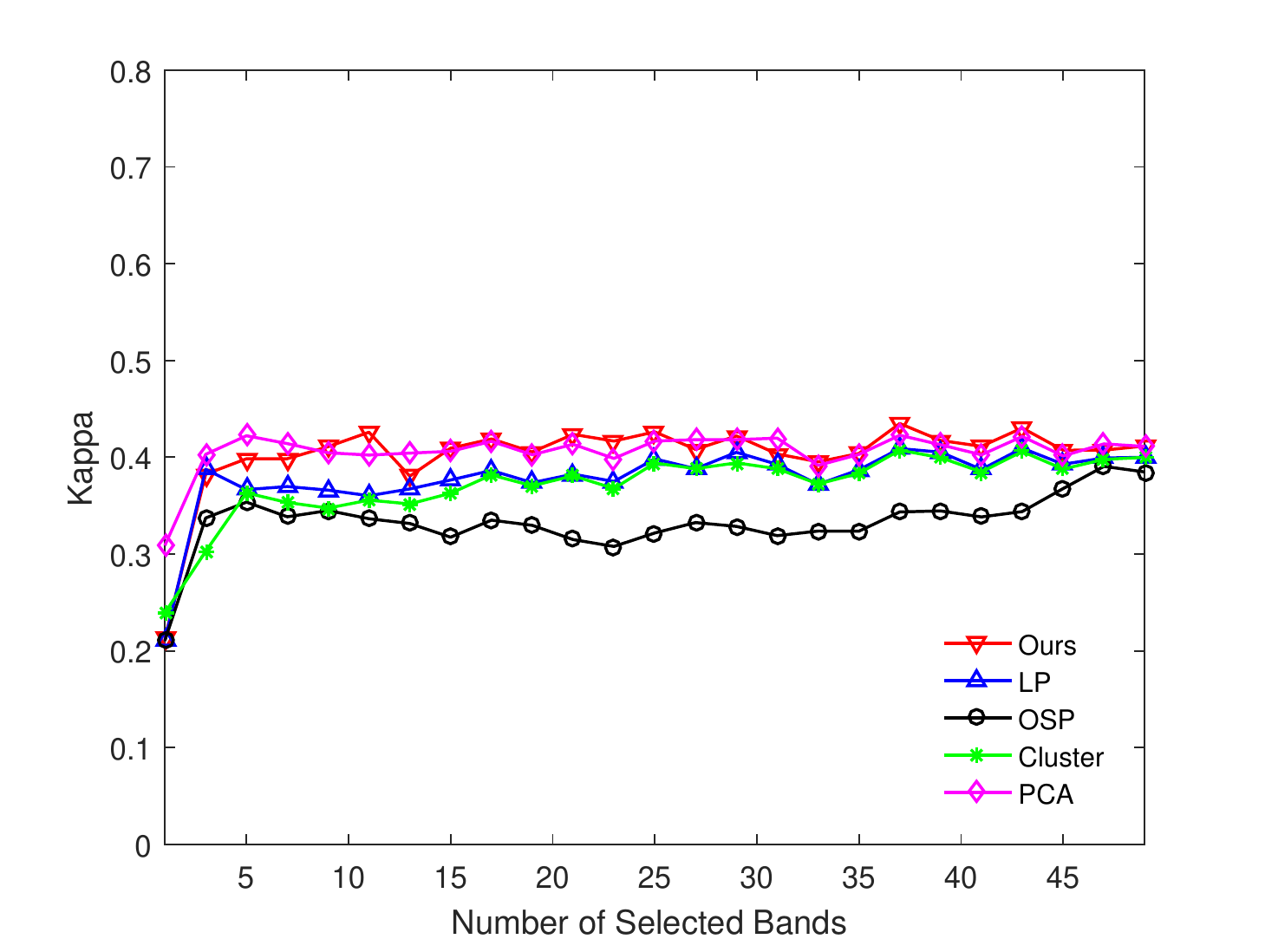} \ \\
 {\small(a) Salinas-A Scene dataset} & {\small(b) Pavia University Scene dataset} & {\small(c) Indian Pines Scene dataset}\\
\end{tabular}
\caption{The Kappa coefficients of different band selection methods using the KNN classifier on the three datasets.
\label{fig:PR-curve}}
\end{center}
\end{figure*}
\begin{figure*}
\begin{center}
\begin{tabular}{@{}c@{}c@{}c@{}c}
\includegraphics[width=0.32\linewidth,height=4.545cm]{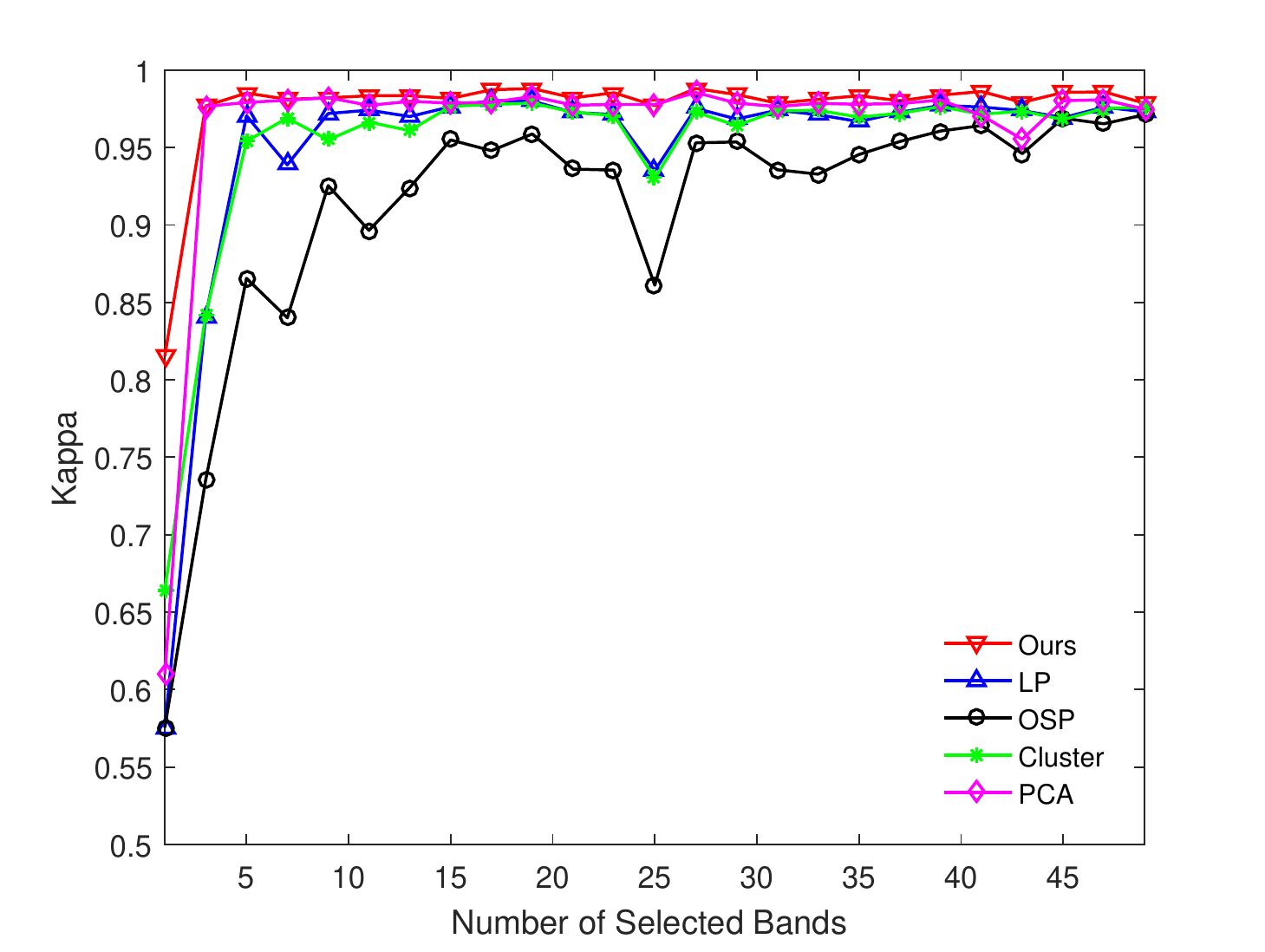} \ &
\includegraphics[width=0.32\linewidth,height=4.545cm]{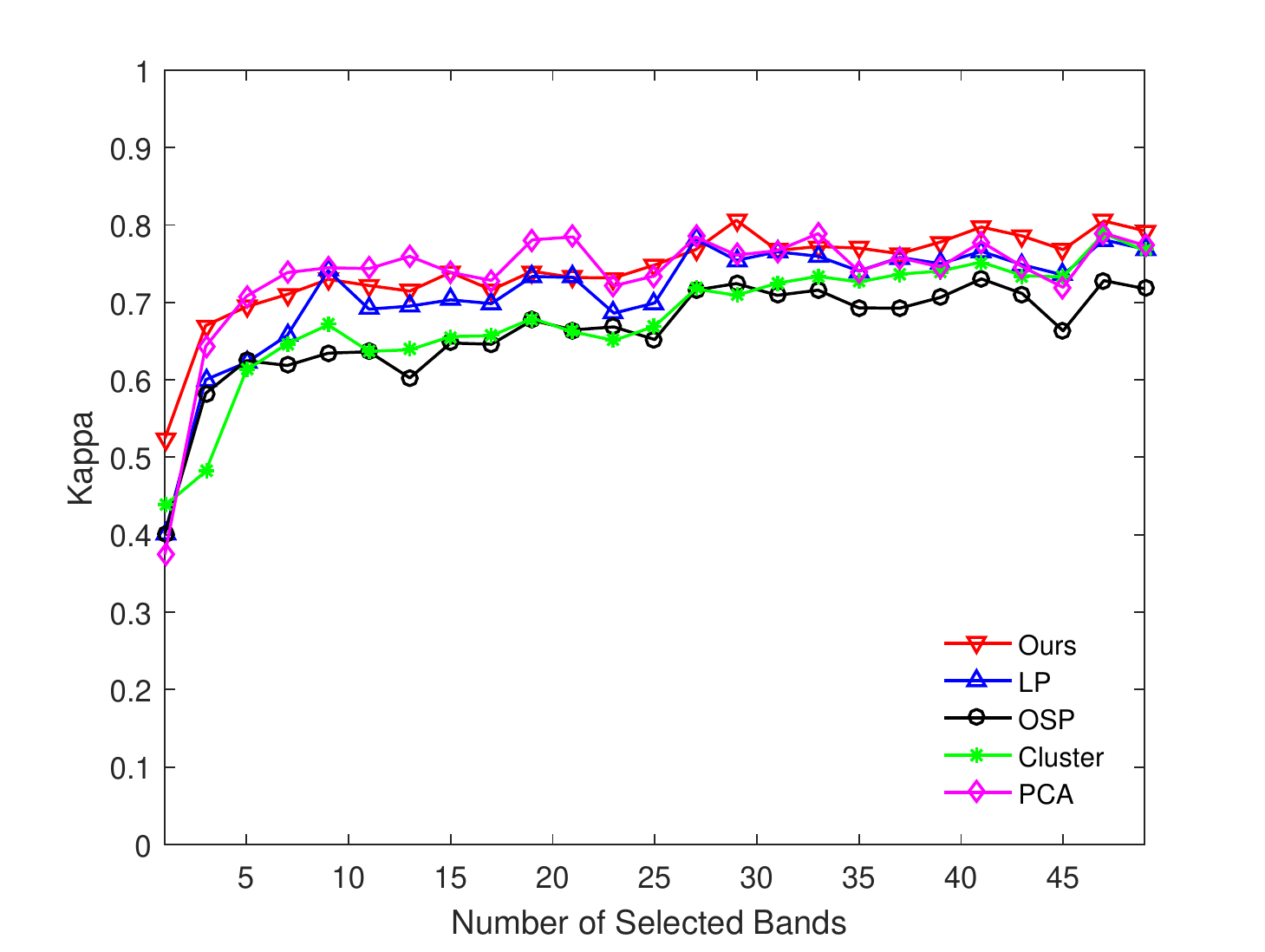} \ &
\includegraphics[width=0.32\linewidth,height=4.545cm]{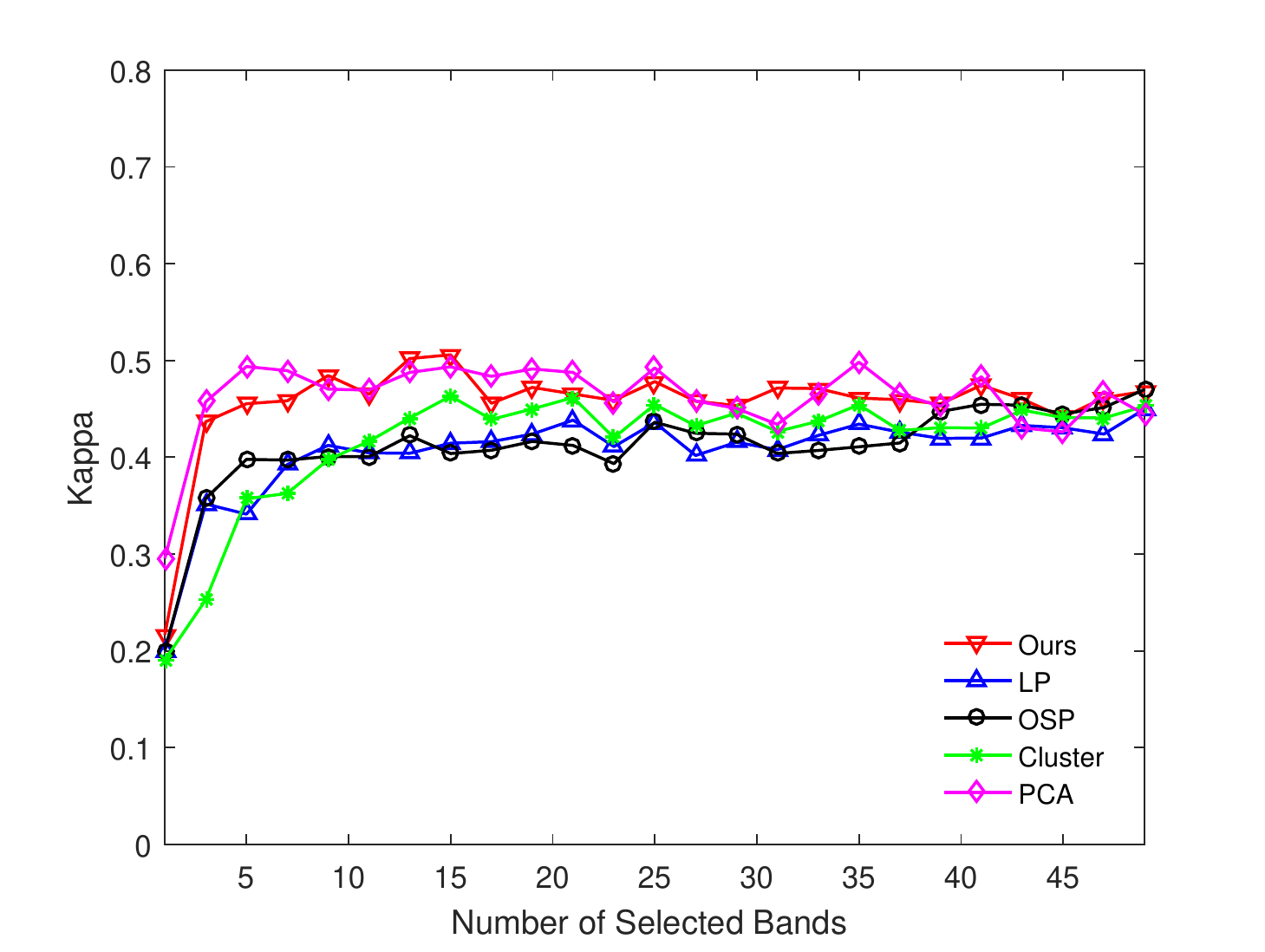} \ \\
 {\small(a) Salinas-A Scene dataset} & {\small(b) Pavia University Scene dataset} & {\small(c) Indian Pines Scene dataset}\\
\end{tabular}
\caption{The Kappa coefficients of different band selection methods using the SVM classifier on the three datasets.
\label{fig:PR-curve}}
\end{center}
\end{figure*}
\begin{figure*}
\begin{center}
\begin{tabular}{@{}c@{}c@{}c@{}c}
\includegraphics[width=0.32\linewidth,height=4.5cm]{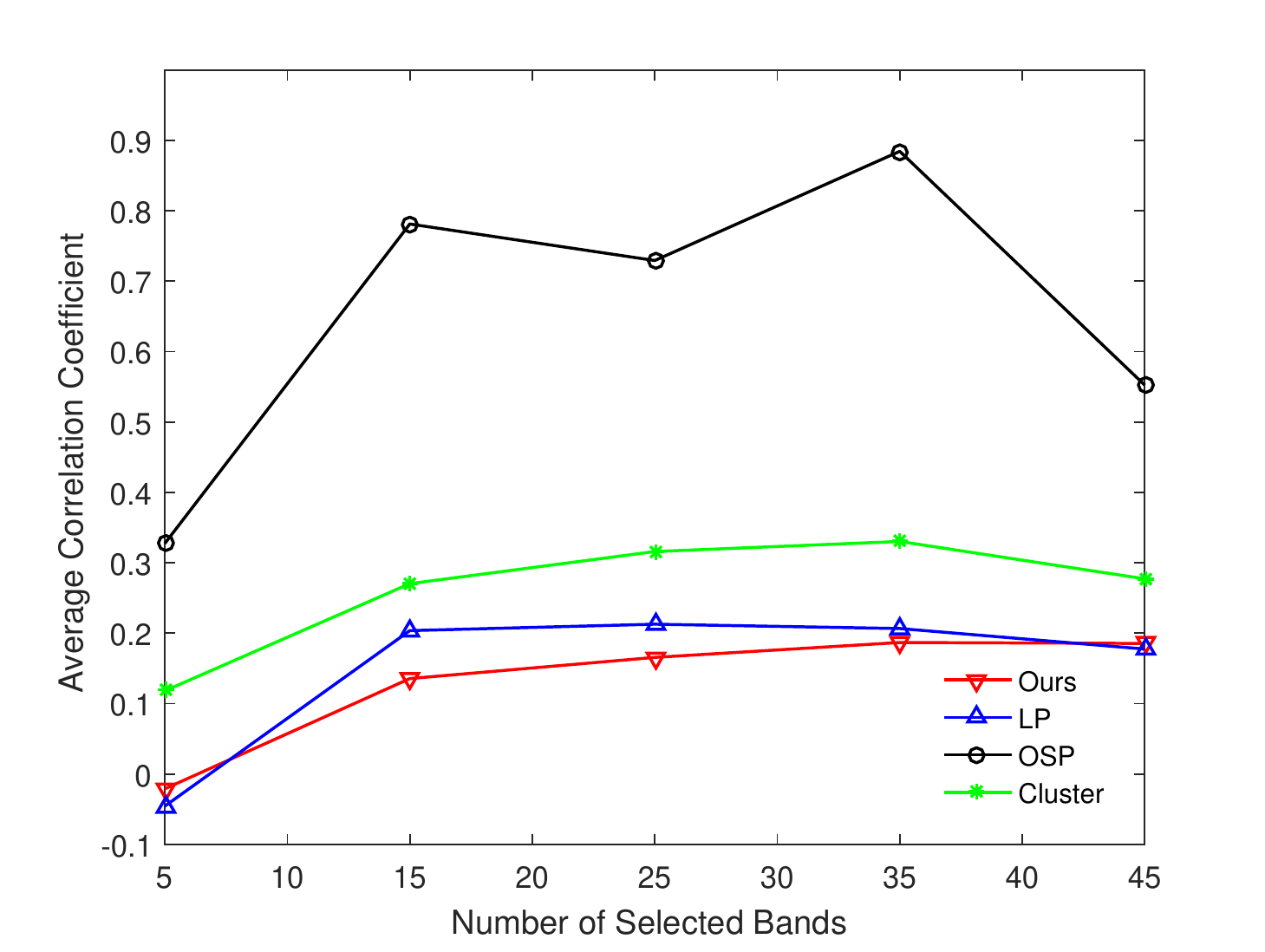} \ &
\includegraphics[width=0.32\linewidth,height=4.5cm]{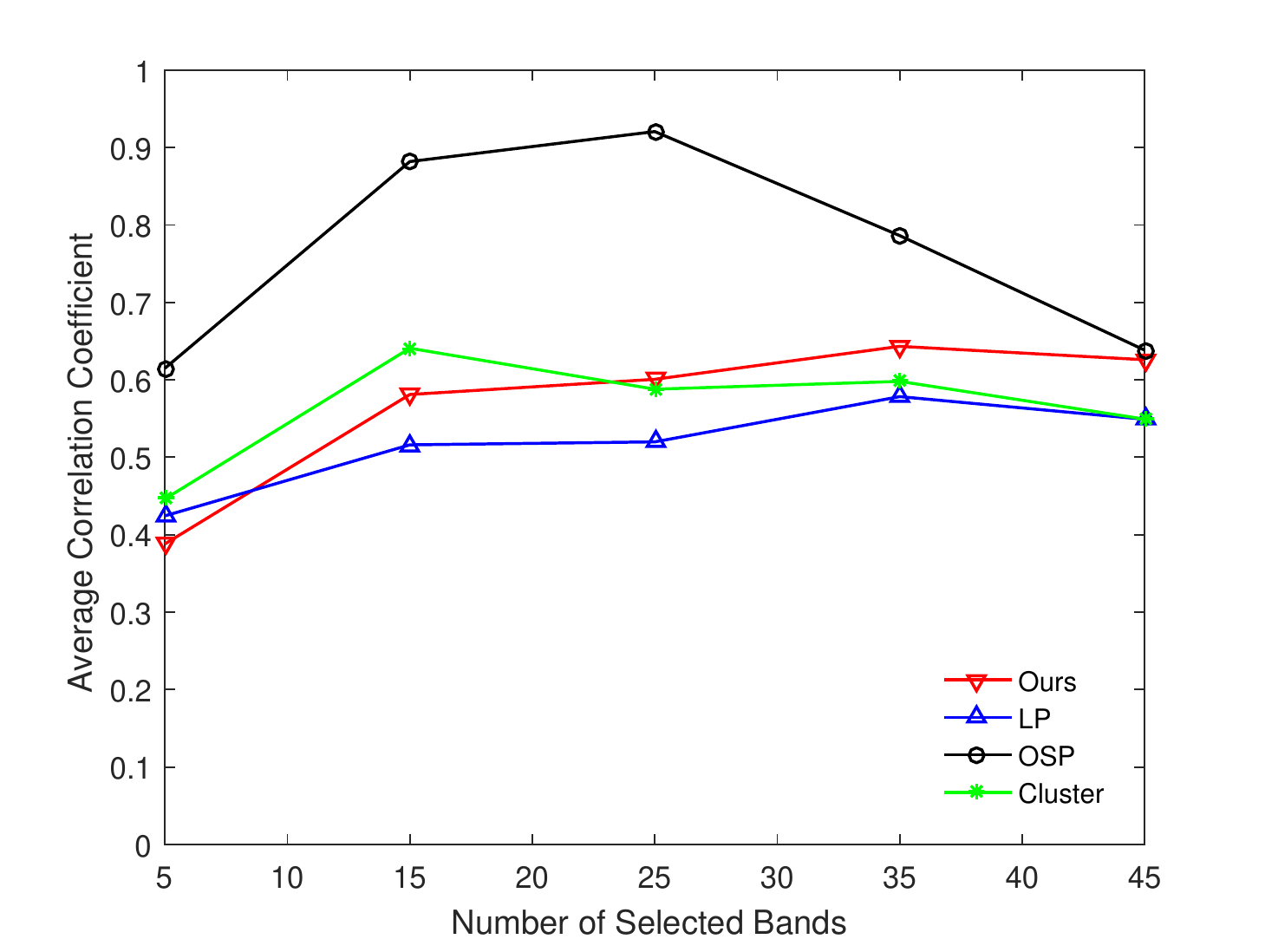} \ &
\includegraphics[width=0.32\linewidth,height=4.5cm]{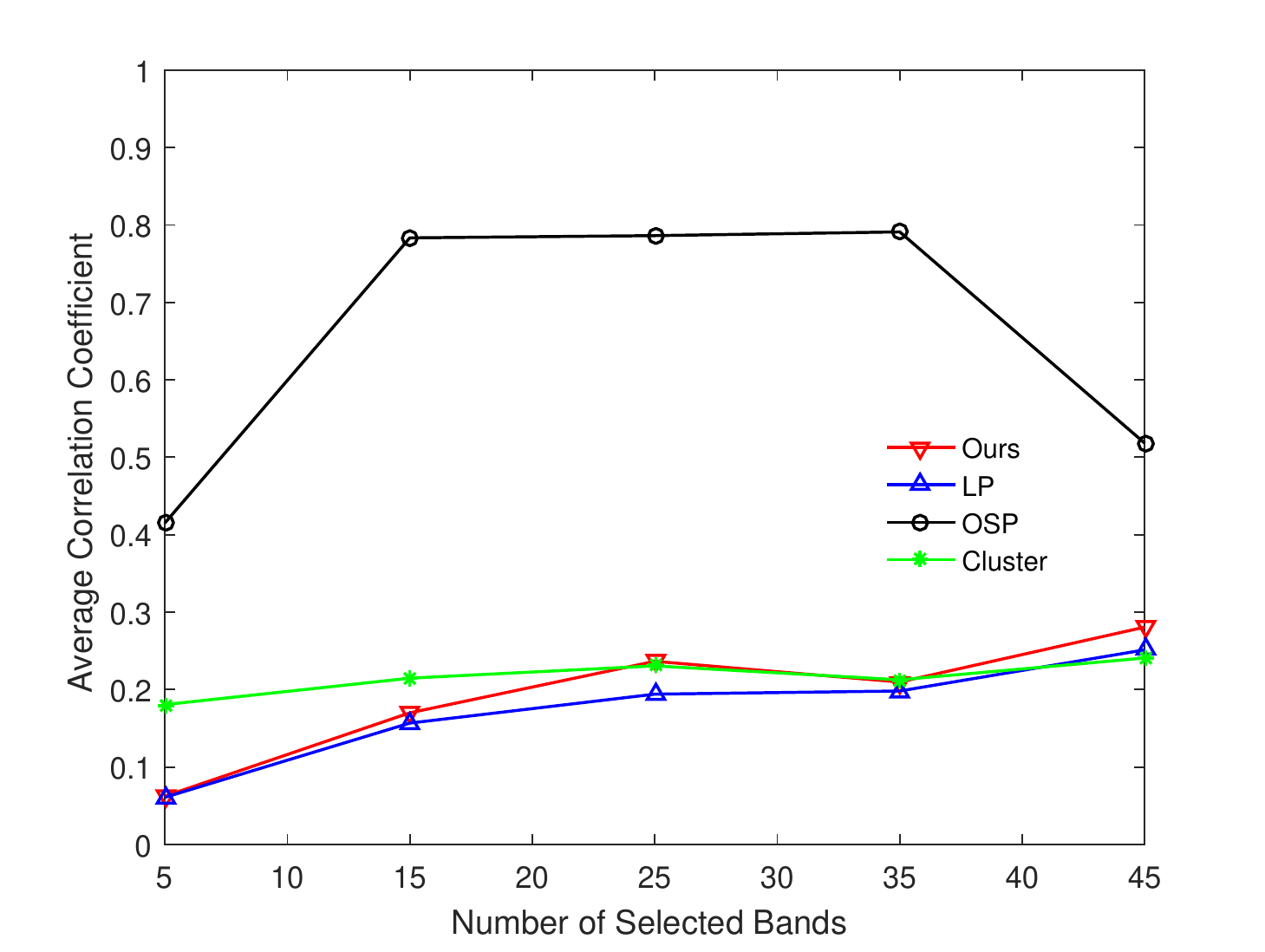} \ \\
 {\small(a) Salinas-A Scene dataset} & {\small(b) Pavia University Scene dataset} & {\small(c) Indian Pines Scene dataset}\\
\end{tabular}
\caption{Comparison of average correlation coefficients with different band selection methods on three datasets.
\label{fig:PR-curve}}
\end{center}
\end{figure*}
\begin{figure*}
\begin{center}
\begin{tabular}{@{}c@{}c@{}c@{}c}
\includegraphics[width=0.32\linewidth,height=4.5cm]{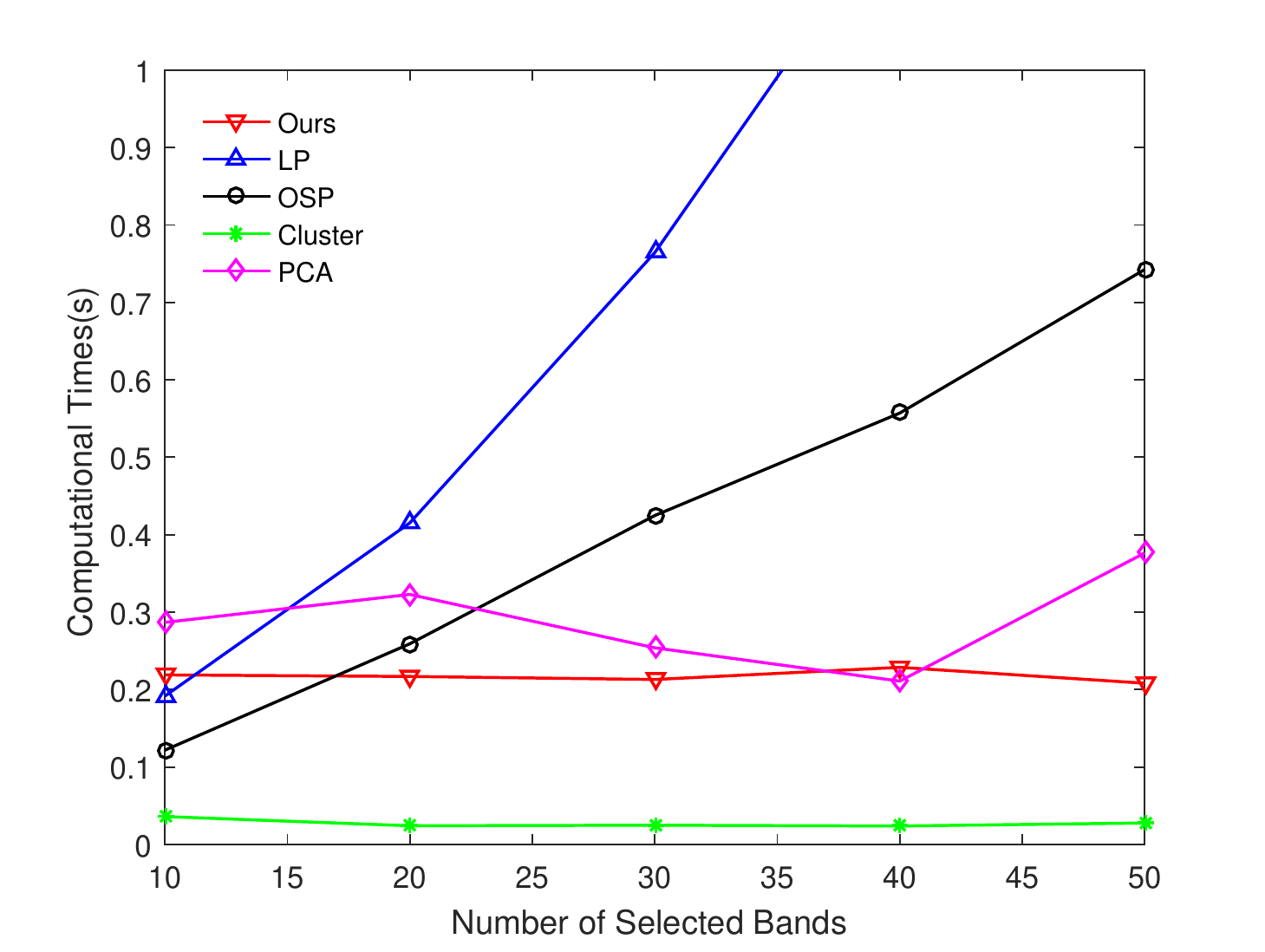} \ &
\includegraphics[width=0.32\linewidth,height=4.5cm]{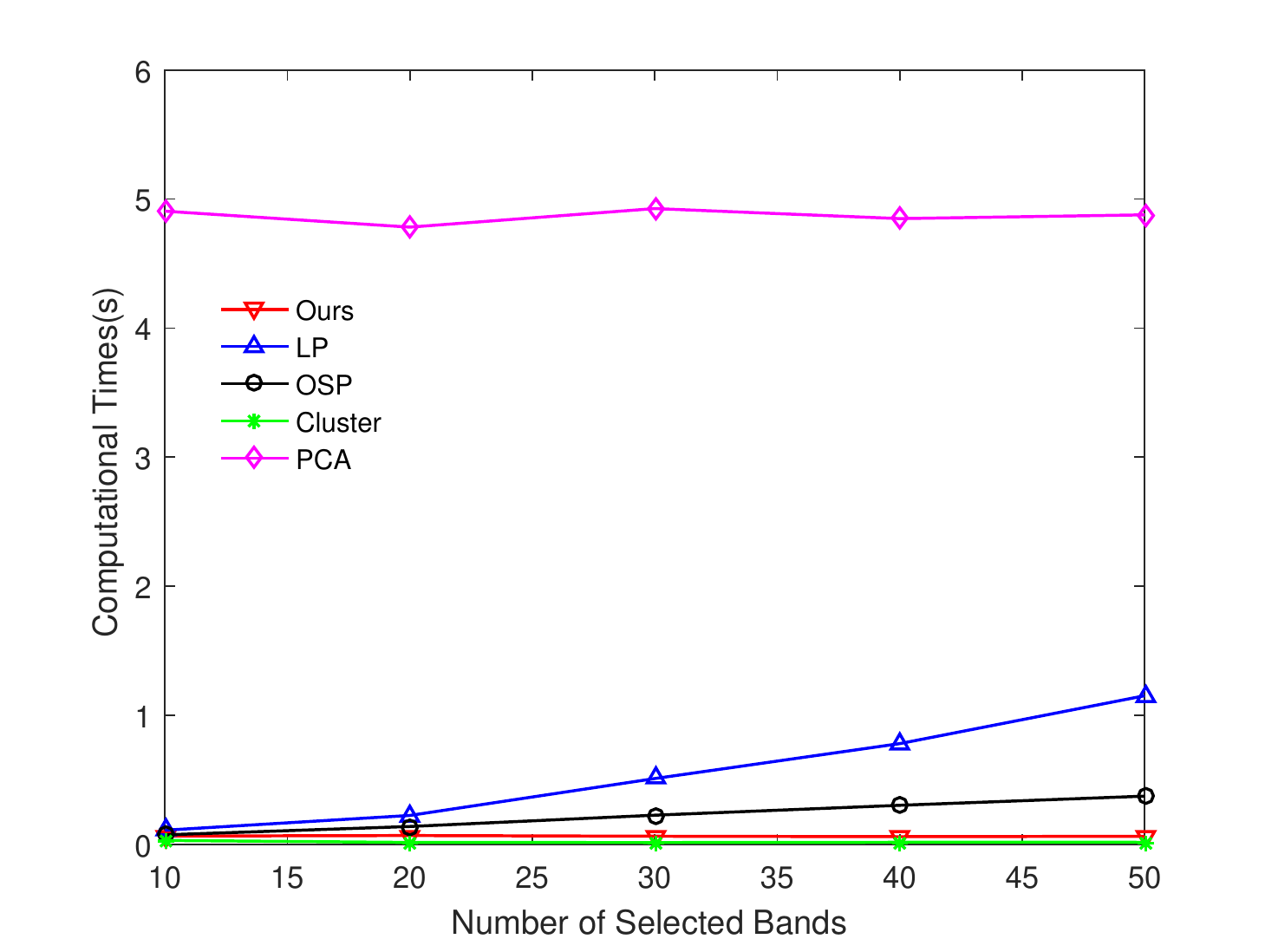} \ &
\includegraphics[width=0.32\linewidth,height=4.5cm]{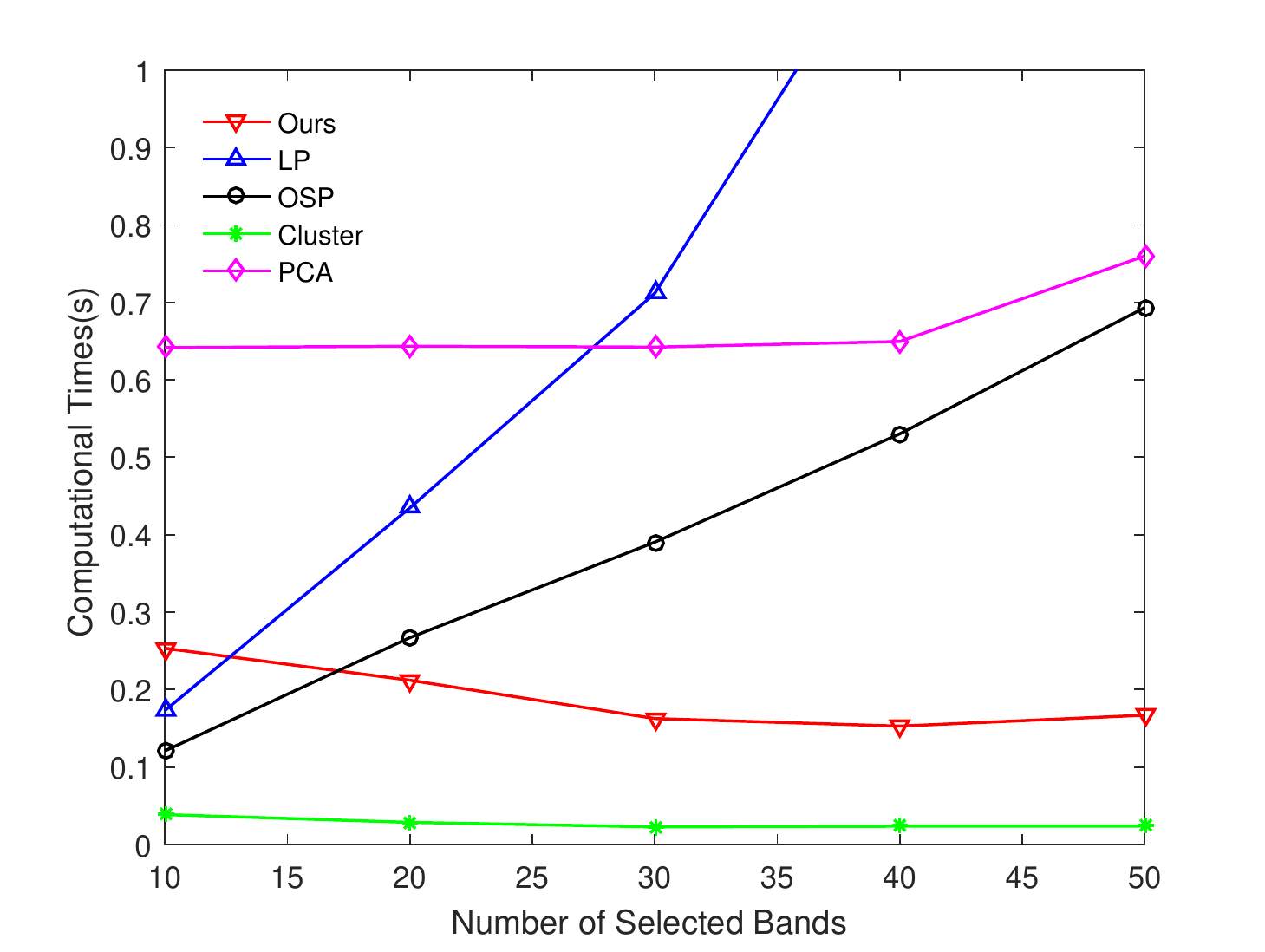} \ \\
 {\small(a) Salinas-A Scene dataset} & {\small(b) Pavia University Scene dataset} & {\small(c) Indian Pines Scene dataset}\\
\end{tabular}
\caption{Comparison of computational time with different band selection methods on three datasets.
\label{fig:PR-curve}}
\end{center}
\end{figure*}
{\flushleft\textbf{Computational Time}}
In order to assess the efficiency of the proposed method, the computational time of each method is also provided.
For fair comparison, all compared methods are implemented in MATLAB2015, and are tested on an Intel Core i5-2400 CPU with 12-GB random access memory.
Figure 13 illustrates the runtime of different band selection methods with varied numbers of bands.
From the results, we can see that the cluster based method takes the least time.
Although our method is not the fastest, it is faster than most of the compared methods.
In addition, our method is not significantly affected by the number of bands.
While the runtime of LP and OSP methods increases quickly as the number of bands ascends.
Therefore, our method has a great advantage in time efficiency.
\section{Conclusion}
In this work, we propose a novel band selection method based on the multi-dictionary sparse representation.
The proposed method is fully unsupervised and consists of three main components: 1) creating multi-dictionaries; 2) optimizing sparse coefficients; 3) computing and sorting weights of desire bands.
Our proposed method not only reduces the dimension of spectral bands, but also preserves the original information of bands.
Extensive experimental results show that the proposed method is very effective and outperforms other competitors.
Of course, the method can be improved.
For example, because of the large number of dictionaries, the amount of calculation increases.
In the future, how to design an effective framework to reduce the number of dictionaries and the amount of calculation is the next work.
\ifCLASSOPTIONcaptionsoff
  \newpage
\fi



\bibliographystyle{IEEEtran}
\bibliography{IEEEabrv,IEEEexample,string,refs}
\end{document}